\documentclass[final,3p,times,fleqn]{elsarticle}

\usepackage[colorlinks,citecolor=red,urlcolor=blue,bookmarks=false,hypertexnames=true]{hyperref} 

\usepackage{graphicx}
\usepackage{caption}
\usepackage{subcaption}
\usepackage{psfrag}
\usepackage{array}
\usepackage{multirow}
\usepackage{booktabs}
\usepackage{threeparttable}
\usepackage{comment}
\usepackage{amsmath}%

\usepackage{xcolor}
\usepackage{amssymb}
\usepackage{amsfonts}%
\usepackage{amssymb}%%
\usepackage{algorithm}
\usepackage{algpseudocode}
\usepackage{setspace}
\usepackage{textcomp}
\usepackage[switch, running]{lineno}
\usepackage{todonotes}
\usepackage{makecell}

\journal{}

\begin{document}
\sloppy
\begin{frontmatter}

\title{Learning CO$_2$ plume migration in faulted reservoirs with Graph Neural Networks}

\author[label1,label3]{Xin Ju\corref{cor1}}%\fnref{label3}}
\address[label1]{Department of Energy Science and Engineering, Stanford University, Stanford, CA, 94305}

%\address[label2]{Atmospheric, Earth, and Energy Division, Lawrence Livermore National Laboratory, Livermore, CA, 94550}
% \address[label2]{Address Two\fnref{label4}}
\address[label3]{TotalEnergies EP Research and Technology USA, Houston, Texas, 77002}

\cortext[cor1]{Corresponding author}
% \fntext[label3]{I also want to inform about\ldots}
% \fntext[label4]{Small city}

\ead{ju1@stanford.edu}
% \ead[url]{author-one-homepage.com}

\author[label3]{Fran\c cois P. Hamon}
\ead{francois.hamon@totalenergies.com}

\author[label1]{Gege Wen}
\ead{gegewen@stanford.edu}

\author[label1]{Rayan Kanfar}
\ead{kanfar@stanford.edu}

\author[label3]{Mauricio Araya-Polo}
\ead{mauricio.araya@totalenergies.com}

\author[label1]{Hamdi A. Tchelepi}
\ead{tchelepi@stanford.edu}

% abstraction
\begin{abstract}
Deep-learning-based surrogate models provide an efficient complement to numerical simulations for subsurface flow problems such as CO$_2$ geological storage.
Accurately capturing the impact of faults on CO$_2$ plume migration remains a challenge for many existing deep learning surrogate models based on Convolutional Neural Networks (CNNs) or Neural Operators.
We address this challenge with a graph-based neural model leveraging recent developments in the field of Graph Neural Networks (GNNs).
Our model combines graph-based convolution Long-Short-Term-Memory (GConvLSTM) with a one-step GNN model, MeshGraphNet (MGN), to operate on complex unstructured meshes and limit temporal error accumulation.
We demonstrate that our approach can accurately predict the temporal evolution of gas saturation and pore pressure in a synthetic reservoir with impermeable faults.
Our results exhibit a better accuracy and a reduced temporal error accumulation compared to the standard MGN model.
We also show the excellent generalizability of our algorithm to mesh configurations, boundary conditions, and heterogeneous permeability fields not included in the training set. 
This work highlights the potential of GNN-based methods to accurately and rapidly model subsurface flow with complex faults and fractures.
\end{abstract}

\begin{keyword}
Graph neural networks, carbon storage, two-phase flow, surrogate model, deep learning
\end{keyword}
\end{frontmatter}
%Introduction
\section{Introduction}
\label{sect:intro}

%Geological carbon storage (GCS) is a promising measure to mitigate the effect of greenhouse gas emissions on climate change. 
%
In geological carbon storage (GCS), large amounts of supercritical $\mathrm{CO}_{2}$ are injected into subsurface reservoirs for permanent storage and must be monitored over very long periods of time to ensure safe and effective storage~\citep{international2020energy, pacala2004stabilization}.
Underground formations often exhibit high degrees of heterogeneity characterized by stratigraphic layering~\citep{mallison2014unstructured} and the presence of faults and fractures~\citep{rinaldi2013modeling, morris2011study}.
These geological features critically impact the migration and the long-term behavior of $\mathrm{CO}_{2}$ plumes. Moreover, faults and fractures could potentially lead to hazards such as induced seismicity or leakage~\citep{castelletto2013multiphysics, cusini2022field}. 
%
%Many studies reported that the presence of fractures and faults in the storage reservoir could cause $\mathrm{CO}_{2}$ leakage and induced seismicity hazards associated with GCS~\citep{castelletto2013multiphysics, cusini2022field}.
%
To ensure the safety and effectiveness of $\mathrm{CO}_{2}$ injection projects, faults and fractures must be accurately modeled in high-fidelity (HF) numerical simulations.
As a result, there is a strong interest in building numerical models based on unstructured polyhedral meshes that conform to the complex geological features of the porous medium~\citep{jenny2002modeling, mlacnik2006sequentially, meng2018local}. % references about PEBI meshes

In addition, managing geological uncertainty in large-scale storage operations requires running a large number of accurate numerical predictions of $\mathrm{CO}_{2}$ plume migration over decades to centuries~\citep{zhu2018bayesian, kitanidis2015persistent, strandli2014co2}. This leads to extremely high computational costs for planning CO$_2$ storage projects with faults. % references about uncertainty qunatifications
%
%HF numerical simulations are often computationally expensive because of the nonlinearity of the governing equations, the large size of the discretized problems, and the cost of solving the nonlinear systems at each time step.
%
Data-driven deep-learning-based (DL) surrogate models for subsurface flow problems have shown great potential to complement HF numerical simulations and reduce the computational burden of uncertainty quantification studies.
Data-driven deep-learning (DL) models rely on data to learn the underlying physics. They approximate the input and output of interest by building statistical models from simulation data generated by HF simulators.
These methods target the minimization of the data loss between prediction fields and label data and can efficiently achieve converged solutions with satisfactory accuracy~\citep{bischof2021multi}.

Previous DL-based models have shown excellent accuracy in predicting flow dynamics~\citep{zhu2018bayesian} and better computational efficiency than HF reservoir simulators~\citep{tang2020deep, tang2021deep}. 
Mo et al. developed a DL surrogate model that integrates an autoregressive model with a convolutional-NNs-based (CNNs) encoder-decoder network to forecast CO$_2$ plume migration in random 2D permeability fields~\citep{mo2019deep}. 
Tang et al.~\citep{tang2020deep, tang2021deep} combined a residual U-Net (R-U-Net) with convLSTM networks to predict the temporal evolution of the saturation and pressure fields in 2D and 3D oil production simulations. 
Their recurrent R-U-Net model was later applied to CO$_2$ storage with coupled flow and geomechanics~\citep{tang2022deep}.
Wen et al.~\citep{wen2021towards} developed an R-U-Net-based surrogate model for CO$_2$ plume migration, where injection durations, injection rates, and injection locations are encoded as channels of input images. 
Recently, Wen et al.~\citep{wen2022u, wen2022accelerating} combined U-Net and Fourier neural operator~\citep{wen2022u} by adding convolutional information in the Fourier layer, which yields significantly improved cost-accuracy trade-off.

However, these existing data-driven surrogate models are limited to Cartesian meshes with simple geometries, which fails to predict in unstructured meshes with stencils that vary in size and shape~\cite {gao2022physics, shukla2022scalable}. 
For example, CNNs are designed for image processing and exploit the Cartesian structure of pixels, implying that these models can only efficiently operate on regular grids~\citep {tang2021deep, tang2022deep}.
This limitation significantly undermines the applicability of these surrogates to field-scale models with complex geological features such as faults and fractures.
%
%  image-based models can only rely on physical fields living on image grids (i.e., cell-centered data), making it challenging to incorporate data defined on faces into the learning procedure.
% %
% %\todo[inline]{MA: second to what? maybe just replace "This second" with "Further, this constraint ..."}
% This constraint further prevents CNNs from capturing nontrivial topological information regarding connectivity and transmissibility, which play an important role in HF numerical simulations and could be used to infuse more physics into the data-driven model.

%subsection 1
%\subsection{Graph Neural Networks}
%\label{sec:GNN}
To overcome these limitations, here we aim to construct a DL surrogate model based on a graph neural network (GNN) that can capture the flow dynamics in realistic subsurface flow problems modeled with complex unstructured meshes conforming to faults and fractures. 
The key feature of GNNs is to represent unstructured simulation data as graphs with nodes and edges, in which the nodes represent cell-centered data (e.g., pore pressure, phase saturation). In contrast, the edges represent cell-to-cell connectivity and face-centered data (e.g., transmissibility, Darcy fluxes).
This is key to enabling the DL surrogate to operate with unstructured-mesh-based simulation data containing complex internal structures.
%
% Multiple GNNs families have been proposed to learn forward dynamics, such as spectral-based methods~\citep{bruna2013spectral, defferrard2016convolutional,chang2021spectral} and spatial GNNs~\citep{gilmer2017neural, hamilton2017inductive, li2018deeper}.
%
Recently, a class of GNNs named message-passing neural networks (MPNNs) has demonstrated its efficiency in learning forward dynamics~\cite{sanchez2020learning, pfaff2020learning, lino2021simulating, shi2022gnn}. 
In MPNNs, a learnable message-passing function is designed to propagate information over the graph through a local aggregation process~\cite{shukla2022scalable}.
Using the local aggregation process works effectively on embedding spaces and helps the model learn better representations. This aggregation operation is pivotal to enable a node to incorporate information from its neighbors, enriching the own representation of the node in the embedding space. As a result, these improved representations produce more accurate predictions and contribute to the potency of MPNNs in handling unstructured graph-based data.
%
%Recently, message-passing neural networks (MPNNs), a type of spatial GNNs, demonstrated their efficiency at learning forward dynamics that involve interaction between different entities~\cite{sanchez2020learning, pfaff2020learning, lino2021simulating, shi2022gnn}.
%
%In the MPNN framework, a learnable message-passing function is designed to propagate information over the graph through a local aggregation (diffusion) process~\cite{shukla2022scalable}. 
%

%
Of particular relevance to this work is the MPNN methodology named MeshGraphNet (MGN) proposed by Pfaff et al. (2020)~\citep{pfaff2020learning}, in which the training graph is constructed from an HF simulation mesh.
The authors demonstrated that by encoding various physical quantities--depending on underlying physics--as edge features of a graph, MGN could be a fast surrogate trained from unstructured HF simulations.
This work also shows that MGN-based surrogate models have the ability to generalize to meshes unseen during training and can capture internal boundaries more accurately than CNN-based models.
Wu et al.~\citep{wu2022learning} applied the MGN architecture to an oil-and-gas problem and developed a hybrid architecture to learn the dynamics of reservoir simulations on Cartesian meshes. The hybrid architecture used a U-Net (a variant of CNNs~\citep{ronneberger2015u}) for the pore pressure and MGN for the phase saturation.  
%
%However, this hybrid setting partially relies on CNNs, thus limiting the model applicability to unstructured subsurface flow simulations. 
%
Notably, these GNN surrogate models all focus on next-state predictions, i.e., they approximate the next state of a physical system from the current state and advance in time in an autoregressive manner. 
However, recent work has shown that next-state models are prone to suffer from substantial temporal error accumulation~\cite{pfaff2020learning, han2022predicting} when autoregressively rolling out for a long period. 
This limitation is problematic to predict $\mathrm{CO}_{2}$ plume behavior, as $\mathrm{CO}_{2}$ operations often require the simulation of multiple decades of injection for a commercial-scale project.
 Therefore, in this work, we introduce a graph-based recurrent neural network to mitigate temporal error accumulation and achieve a reliable long-term prediction. 

%
% In this study, we present a recurrent GNN to learn the temporal dynamics of subsurface flow. %The network architecture is inspired by the combination of MeshGraphNet, a next-state predictor, and a graph-based recurrent model. 
%
The proposed recurrent GNN framework includes (1) a modified MGN that encodes and processes the current physical state into embedding spaces of the entire graph and (2) a graph-based recurrent convLSTM (GConvLSTM)~\cite{seo2018structured} model that predicts the next state based on the embeddings computed by MGN and on recurrent memories from past states. 
%
%Specifically, we utilize a graph-based convLSTM~\cite{seo2018structured} (GConvLSTM) to capture the temporal evolution of the physical state. 
%
In comparison to the original next-state MGN predictor, the proposed algorithm, referred to as MeshGraphNet-Long Short-Term Memory (MGN-LSTM), can better mitigate temporal error accumulation and significantly improves the performance for the long-term prediction of CO$_2$ plume behavior.
Our MGN-LSTM algorithm can accurately approximate HF simulations on unstructured meshes and is generalizable to meshes, boundary conditions, and permeability fields unseen during training.
Our main contributions include: (1) using GNN to overcome the current limitations of previous surrogate models to handle complex simulation data on unstructured meshes in the context of CO$_2$ geological storage in faulted reservoirs; (2) introducing the accurate MGN-LSTM architecture reducing temporal error accumulation; and (3) demonstrating the generalizability of MGN-LSTM to unseen meshes and boundary conditions as well as its stable extrapolation to future states.
%and (4) infusing underlying physics via augmenting input features to obtain a physics-based message passing layer. 
%\todo[inline]{MA: the GNN vs CNN angle is not present in this paragraph}
%\subsection{Key contributions}
%\label{sec:review}

%
This paper proceeds as follows.
In Section~\ref{sec:problem_statement}, we introduce the two-phase flow equations applicable to CO$_2$ geological storage.
Section~\ref{sec:gnn_model} describes the proposed surrogate model (MGN-LSTM) and associated data-processing and training procedures.
In Section~\ref{sec:results}, the MGN-LSTM is used to predict saturation and pore pressure fields in two-phase flow (CO$_2$-brine) problems.
%
%Each high-fidelity simulation (HF) is performed with varying mesh, well location, and permeability.
%
%We verify the applicability of the surrogate model to learn physics embedded in the HF simulation results.
%
%A detailed assessment of surrogate model accuracy is presented.
Section~\ref{sec:comparison_with_standard_mgn} demonstrates the improved accuracy of MGN-LSTM compared to the standard MGN algorithm.
Section~\ref{sec:conl} concludes the work and suggests future research directions.

% Methodology 
\section{Problem statement}
\label{sec:problem_statement}

\subsection{Governing equations of CO$_2$-brine flow}
\label{sec:pde}

In this work, we consider miscible two-phase (gas and aqueous) two-component (H$_2$O and CO$_2$) flow in a compressible porous medium.
We employ a 2D domain in the $x-y$ plane for simplicity, but the model and algorithms presented here can be extended to 3D.
The H$_2$O component is only present in the aqueous phase, while the CO$_2$ component can be present in both the aqueous and the gas phases.
We denote the aqueous and the gas phases using the subscripts $a$ and $g$, respectively.
The mass conservation of each component reads:
\begin{equation}
  \frac{\partial}{\partial t} \left( \phi \sum_{\ell = 1}^{2} x_{c\ell} \rho_\ell s_\ell \right) + \nabla \cdot \left( \sum_{\ell = 1}^{2} x_{c\ell} \rho_\ell \boldsymbol{v}_\ell \right) +  \sum_{\ell = 1}^{2} x_{c\ell} \rho_\ell q_\ell = 0, \qquad c = \{ \text{H}_2\text{O}, \text{CO}_2 \},
  \label{eq:fluid-mass-balance}
\end{equation}
where $\phi$ is the porosity, $x_{c\ell}$ is the mass fraction of component $c$ in phase $\ell$, $\rho_\ell$ is the density of phase $\ell$, $s_\ell$ is the saturation of phase $\ell$, $\boldsymbol{v}_\ell$ is the Darcy velocity of phase $\ell$, and $q_\ell$ is the source flux for phase $\ell$.
Using the multiphase extension of Darcy's law, we write that the Darcy velocity is proportional to the gradient of the pressure:
\begin{equation}
  \boldsymbol{v}_\ell = -\frac{k_{r\ell}}{\mu_\ell} \bar{\mathbf{k}} \cdot \nabla p_\ell, \qquad \ell = \{ a, g \},
  \label{eq:darcy-eq}
\end{equation}
where $k_{r\ell}$ is the relative permeability of phase $\ell$, $\mu_\ell$ is the viscosity of phase $\ell$, $\bar{\mathbf{k}}$ is the permeability tensor, and $p_{\ell}$ is the phase pressure.
In this work, we assume that the permeability tensor is diagonal with an equal value for each entry.
The system is closed with the following constraints
\begin{align}
  &s_g + s_a = 1,  \qquad &\text{(saturation constraint)} \\
  &p_g - p_a = p_c(s_g), \qquad &\text{(capillary pressure constraint)} \\
  &x_{\text{H}_2\text{O},\ell} + x_{\text{CO}_2,\ell} = 1, \qquad \ell \in \{ a, g \}, \qquad &\text{(component fraction constraints)}
\end{align}
as well as standard thermodynamics constraints on fugacities.
The partitioning of the mass of the CO$_2$ component between the gas phase and the aqueous phase is determined as a function of pressure, temperature, and salinity using the model of Duan and Sun \cite{duan03}.
The gas phase densities and viscosities are computed using the Span-Wagner~\cite{span96} and Fenghour-Wakeham~\cite{fenghour98} correlations,
respectively, while the brine properties are obtained using the methodology of Phillips et al. \cite{phillips81}.
The relative permeabilities are computed with the Brooks-Corey model as $k_{rg}(s_g) = 0.95 ( s_g / s_{g,max} )^2$ and $k_{ra}(s_a) = ( (s_a - s_{a,min}) / (1-s_{a,min}) )^6$.
The capillary pressure $p_c$ is computed from $s_g$ using the Leverett J-function relationship.
The domain is initially saturated with brine, with an initial pressure of 10 MPa and an initial temperature of 143.76 degrees Celsius.
We use analytical (Carter-Tracy) aquifer boundary conditions.
A well injects pure supercritical CO$_2$ at a rate of 0.058 kg/s for 950 days, assuming a storage reservoir with unit meter thickness.
%\todo[inline]{MA: remember to fill in X}

\subsection{Finite-volume discretization on unstructured polygonal meshes}

Considering a mesh with $n_{C}$ cells, the system of equations (\ref{eq:fluid-mass-balance})-(\ref{eq:darcy-eq}) is discretized with a cell-centered, fully implicit (backward-Euler) finite-volume scheme based on a two-point flux approximation (TPFA) and single-point upstream weighting.
The primary variables are chosen to be the gas (CO$_2$-rich) phase pressure $p = p_g$, the overall component densities $\rho_{\text{H}_2\text{O}}$, and $\rho_{\text{CO}_2}$, where an overall component density represents the mass of a given component per unit volume of mixture.
The primary variables
can be related to the variables of equation (\ref{eq:fluid-mass-balance}) using the formulas given in \cite{voskov2012comparison}.
At each time step, the nonlinear system of discretized equations is solved with Newton's method with damping to update all the primary variables in a fully coupled fashion.
All the simulations are performed with GEOS, an open-source multiphysics simulator targeting geological carbon storage and other subsurface energy systems \cite{bui2021multigrid,t2022deformation,cusini2022field}.

\begin{figure}[htbp]
  \centering
  \includegraphics[width=6.4in,height=2.7in]{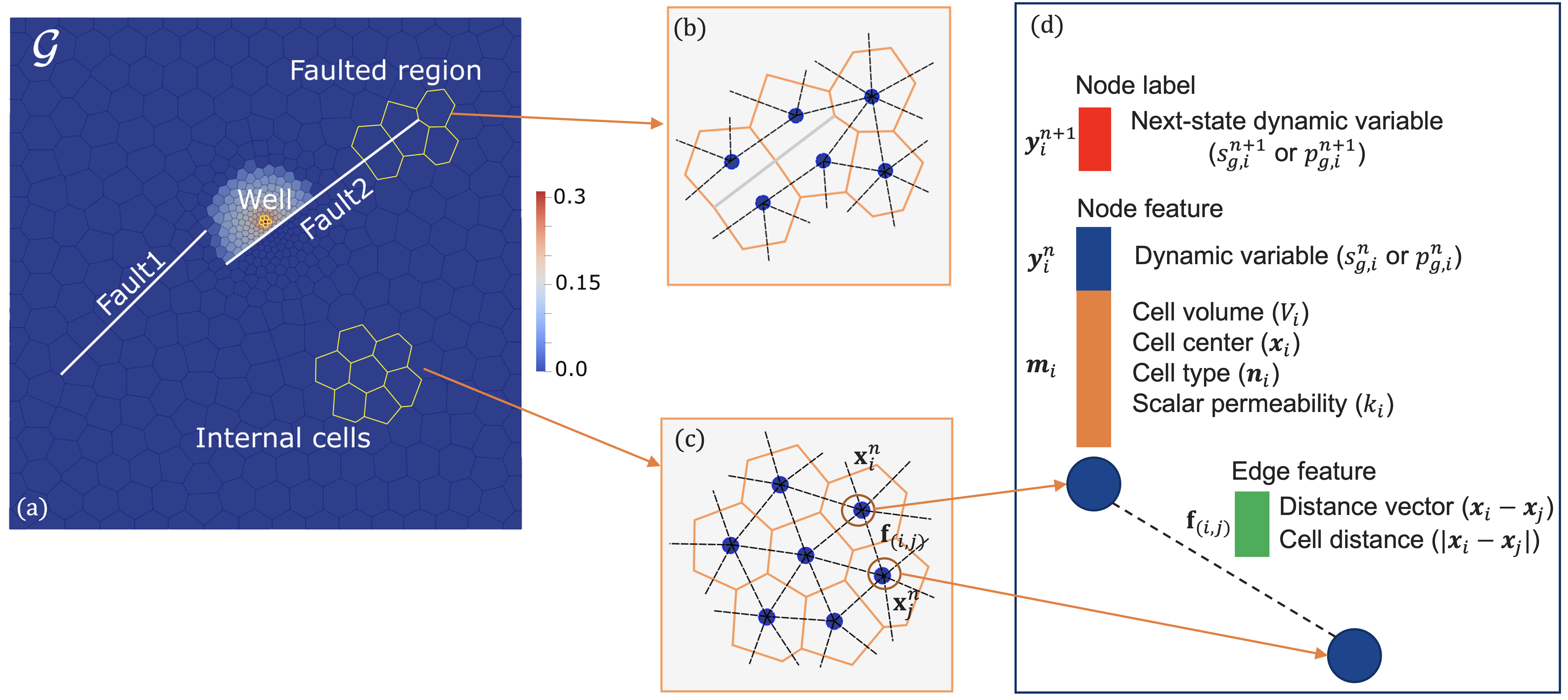}
   \caption{Construction of the graph representation from an unstructured mesh: (a) an example mesh with two impermeable faults and one injector; highlighted areas are zoomed-in sections of (b)  cells along fault line \#2; (c) internal cells and their graph representations; and (d) input node features and edge features as well as node output (node label).}
  \label{fig:graph_construction}
\end{figure}

Unstructured polygonal meshes are well suited to represent complex faults and to perform local spatial refinement around the wells.
In this work, we focus on the class of perpendicular bisector (PEBI) grids \cite{palagi1994use,jenny2002modeling,mlacnik2006sequentially,meng2018local,borio2021hybrid} generated with MRST (Matlab Reservoir Simulation Toolbox)~\cite{lie2016introduction} to mesh a 1 km x 1 km x 1 m domain containing an injector well and two straight impermeable faults in a conforming fashion (see an example mesh shown in Figs.~\ref{fig:graph_construction}(a)). 
As PEBI meshes are orthogonal by construction, the flow simulations can be performed with the TPFA finite-volume scheme without compromising the accuracy of the solution.
%\todo[inline]{Isaac: We need to mention the relative permeability equation used in this chapter. it will later be referenced when introducing the injecting physics part?}
% Recurrent GNN
\section{Recurrent GNN surrogate model}
\label{sec:gnn_model}

In this section, we first describe the graph representation of unstructured mesh-based simulation data. Then, we introduce the key aspects of the deep-learning-based surrogate model for two-phase subsurface flow problems, including the learning task, model architecture, training procedure, and data preparation.

\subsection{Input and output graph representation}
\label{subsec:graph_rep}

To leverage the capabilities of graph-based machine learning, we represent the unstructured mesh-based input and output data at a given time as a directed graph with properties.
The graph is denoted by $\mathcal{G} = (\mathcal{V}, \mathcal{E})$ where $\mathcal{V} $ and $ \mathcal{E} $ are node and edge sets, respectively.
As shown in Fig.~\ref{fig:graph_construction}(a), each mesh cell $i$ is represented by a graph node $i \in \mathcal{V}$.
Two adjacent cells $i$ and $j$ connected with non-zero transmissibility are represented by two graph edges, with edge $(i,j) \in \mathcal{E}$ pointing to node $j$ and edge $(j,i)$ pointing to node $i$.
The transmissibility $T_{ij}$ at each connection between cells $i$ and $j$ is computed as a function of grid geometry and rock permeability as explained in \cite{lie2016introduction, lie2021advanced}, and is equal to zero when two cells are separated by a fault.
%\todo[inline]{MA: adaptive mesh/graph as future work? maybe we can mentioned here so some else give it a shot}
For a given set of mesh and permeability, the graph structure remains
fixed for all time steps.
The properties associated with node $i$ at time $t_n$ are termed as node feature, $\mathbf{x}_i^n : \mathcal{V} \rightarrow \mathbb{R}^{n_N}$. 
The properties corresponding to edge $(i, j)$ are referred to as edge feature, $\mathbf{f}_{(i, j)}: \mathcal{E} \rightarrow \mathbb{R}^{n_E}$ and are independent of time. 
The dimensions of the node and edge features, $n_N$ and $n_E$, are specified below.

As illustrated in Fig.~\ref{fig:graph_construction}(d), the node feature $\mathbf{x}_i^n$ consists of a combination of dynamic variables at time $t_n$, $y_i^{n} \in \mathbb{R}$, and static model features $\mathbf{m}_i \in \mathbb{R}^{n_M}$, where $n_M$ is the dimension of the static model features.
Specifically, $y_i^{n}$ denotes the state variable whose dynamics are learned by the neural network, such as phase saturation, $s_{g,i}$, or pore pressure, $p_{g,i}$.
%
%\todo[inline]{MA: another potential future path, joint learning or a multi-output net.}
Note that in this work, for each dynamic variable ($s_g$ or $p_g$), we train one individual prediction model.
The static model parameters $\mathbf{m}_i$ include the scalar permeability $k_i$ (one dimension), the cell volume $V_i$ (one dimension), the cell center coordinates $\boldsymbol{x}_i$ (two dimensions), and the cell type $\boldsymbol{n}_i$ (a one-hot vector of dimension 4), such that the dimension of the model parameters is $n_M = 1 + 1 + 2 + 4 = 8$ and the dimension of the node features is $n_N = n_M + 1 = 9$.
%\todo[inline]{what does "such that $n_M = 4$ and $n_N = n_M + 1 = 5$" mean? }
%
As shown in Figs.~\ref{fig:graph_construction} (b) to (c), the cell type is used to identify cells playing a key role in the simulation, such as the cells along the faults and the cells where source terms (well) or boundary conditions are imposed. 
The fourth cell type includes the remaining cells (not along faults, and not where source and boundary conditions are imposed).
%Specifically, $\boldsymbol{n}_i$ is a one-hot vector of size 4.
%
The edge feature $\mathbf{f}_{(i,j)}$ of edge $(i, j)$ is constructed to enrich the graph connectivity information with the (signed) distance between cell centers, $\boldsymbol{x}_j – \boldsymbol{x}_i \in \mathbb{R}^2$, (see Fig.~\ref{fig:graph_construction}(d)) and its absolute value, such that $n_E = 3$.
%
%As we treat the two-phase flow problem as an Eulerian \cite{anderson1995computational} system, the graph $\mathcal{G}$ remains fixed for all time steps.  
The schematic expressions of node and edge features used in this study are given in Fig.~\ref{fig:graph_construction}(d)). We explored four configurations of node and edge inputs; the corresponding variables are detailed in Table~\ref{tab:node_edge_feature}.

%\todo[inline]{MA: missing table reference}
%\todo[inline]{Isaac: Given our graph is quite informative, I swapped the table with figure to supplement explaining the features. 

%I have put the table into the section where we explain the effect of infusing physics on the model performance. This is because i have redesigning the table by adding three other rows representing new cases.
%}

\subsection{MeshGraphNet-Long Short-Term Memory (MGN-LSTM)}
\label{subsec:learning_taks}

%\todo[inline]{MA: do we need to mention here that the inference on the trained model produces a surrogate?}
The proposed MGN-LSTM model is designed to learn the spatio-temporal evolution of the selected dynamic variable (pressure or CO$_2$ saturation) of the two-phase flow problem defined in Section~\ref{sec:pde}. 
Given the initial state $\mathbf{Y}^{0} = [y^0_1, \dots, y^0_{n_{C}}]^T$ and the static model features $\mathbf{M} = [(\mathbf{m}_1)^T, \dots, (\mathbf{m}_{n_{C}})^T]^T$ defined at the nodes of $\mathcal{G}$, and given the features at all edges $\mathbf{F}$, we compute the sequence of dynamic variables $(\hat{\mathbf{Y}}^{1}, \dots ,\hat{\mathbf{Y}}^{n_{T}})$ in an autoregressive way as follows:
\begin{align}
     & \hat{\mathbf{Y}}^{0} = \mathbf{Y}^{0}, \\
     &\hat{\mathbf{Y}}^{n+1} = f_{\text{MGN-LSTM}, \theta} ( \mathcal{G}, \hat{\mathbf{Y}}^{n}, \mathbf{M}, \mathbf{F}), \qquad n \in \{1, \dots, n_{T}\}, 
    \label{eq-RMGN} 
\end{align}
where  $f_{\text{MGN-LSTM}, \theta}$ is the surrogate model parameterized by the model weights $\theta$. 
The number of cells in the mesh is denoted by $n_C$ and the number of temporal snapshots by $n_T$.
Once the models are trained, the inference is identical to the training process of Eq.~(\ref{eq-RMGN}) and  produces a surrogate model with all the learnable parameters being fixed. 
The training process will be discussed in detail in the next section.

\begin{figure}[htbp]
  \centering
  \includegraphics[trim={0cm 0cm 0cm 0cm},clip, scale=0.72]{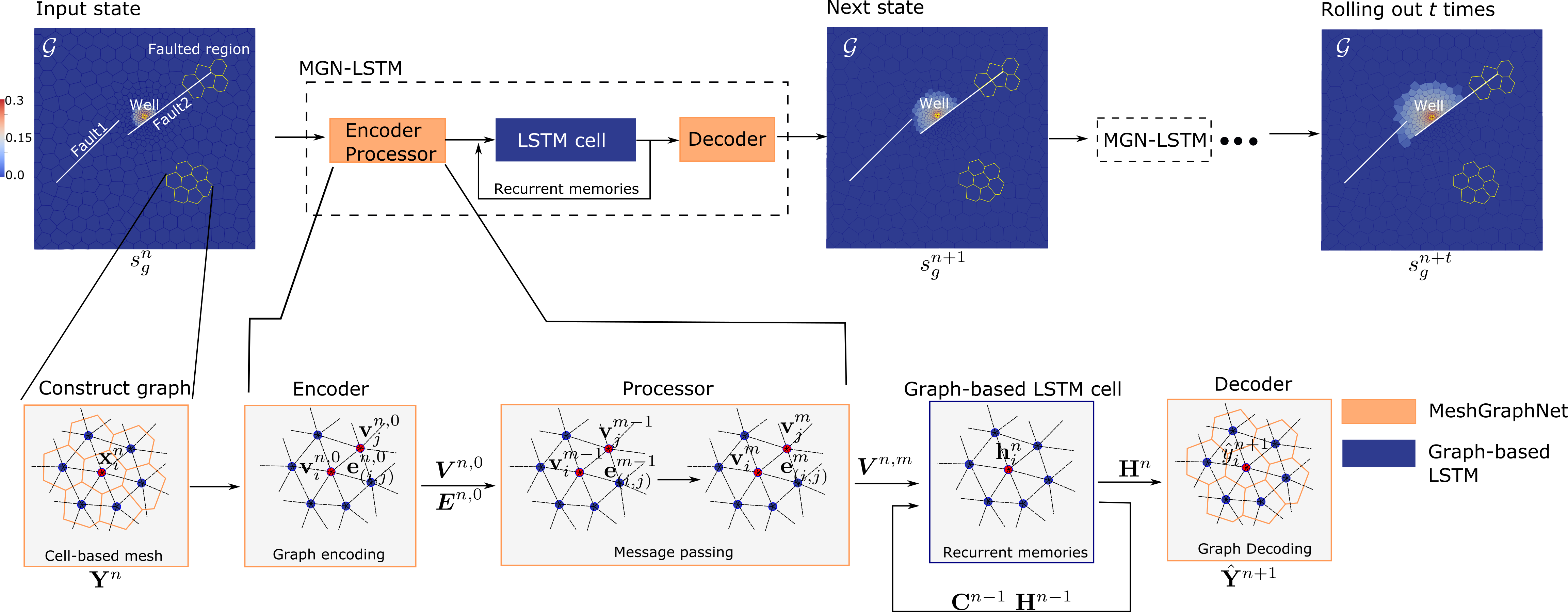}
   \caption{Workflow schematic for the recurrent GNN model proposed in this article, referred to as MGN-LSTM. Given a domain in which supercritical CO$_2$ is injected in the center, the goal of MGN-LSTM is to learn a sequence of dynamic variables. The domain is discretized using an unstructured mesh that conforms to two fault lines and an injection well. A GNN framework based on MGN is used to predict physical quantities from time step $n$ to time step $n+1$ until the end of the simulation. A graph-based recurrent neural network framework based on Graph ConvLSTM ~\cite{seo2018structured} is employed to better capture and represent temporal dynamics.}
  \label{fig:RMGN}
\end{figure}

%\subsection{Architecture of Recurrent MeshgraphNet (RMGN)}

In MGN-LSTM, the temporal evolution of the dynamic variable is captured by learning directly on the graph using latent node and edge features derived from the physical node and edge features reviewed in Section \ref{subsec:graph_rep} as in the recent work of Pfaff et al. \cite{pfaff2020learning}.
Specifically, our algorithm, sketched in Fig.~\ref{fig:RMGN} and described in the next sections, combines the encoder-processor-decoder procedure of MGN \cite{pfaff2020learning} with a graph-based sequence model named Graph ConvLSTM \cite{seo2018structured} to perform the learning task.
Let us consider a one-step prediction, i.e., the computation of $\hat{\mathbf{Y}}^{n+1}$ using $\hat{\mathbf{Y}}^{n}$ in Eq.~(\ref{eq-RMGN}).
First, the encoder-processor steps detailed in Section~\ref{subsec:encoder_processor} map the physical node/edge features to the latent node/edge features.
Then, as described in Section~\ref{subsec:graph_convLSTM}, the latent space variables are used as input to the Graph ConvLSTM algorithm, which aims at retaining spatial-temporal information encoded in the recurrent memories.
Finally, the output of Graph ConvLSTM is decoded and mapped to the physical space using the procedure of Section~\ref{subsec:decoder}.
At that point, the one-step prediction of the dynamic variable is complete.

In the following subsections, we define the main components of the proposed architecture.

\subsubsection{Encoder and processor}
\label{subsec:encoder_processor}

The encoder is the first step of the prediction in MGN-LSTM.
Using the graph representation of Section~\ref{subsec:graph_rep}, we compute the initial latent feature vectors at time $t_n$, 
$\mathbf{v}_{i}^{n, 0} \in \mathbb{R}^{n_H}$ and $\mathbf{e}_{i}^{n,0} \in \mathbb{R}^{n_H}$ from the physical feature vectors, $\mathbf{x}_{i}^{n}$ and $\mathbf{f}_{(i,j)}$.
The hyperparameter $n_H$ denotes the size of the latent vectors. 
The computation of the latent vectors is done using the node and edge multilayer perceptrons (MLPs), denoted respectively by $\operatorname{mlp}_{v}^{0}$ and $\operatorname{mlp}_{e}^{0}$, as follows:
%After representing the simulation as an input graph, $\mathcal{G}^{t}$, we use the deep encoder to independently embed the features for each node, $i$, and edge, $(i,j)$, into latent vectors. This is done with the node and edge multilayer perceptrons 
%(MLPs), which can be represented by equ.\ref{encoder1} and \ref{encoder2}, respectively:

\begin{equation}
\begin{aligned}
\mathbf{v}_{i}^{n, 0} &=\operatorname{mlp}_{v}^{0}\left(\mathbf{x}_{i}^{n}\right), \\
\mathbf{e}_{(i,j)}^{n, 0} &=\operatorname{mlp}_{e}^{0}\left(\mathbf{f}_{(i,j)}\right).
\end{aligned}
\label{encoder}
\end{equation}
%
%The expressions for $\operatorname{mlp}_{v}^{0}$ and $\operatorname{mlp}_{e}^{0}$ can be found in Appendix X. 
%
%\todo[inline]{MA: forget not to fill in X}

The graph equipped with the initial latent node and edge features computed by the encoder at time $t$ in Eq.~(\ref{encoder}) is the input to the processor.
The processor consists of $m$ message-passing steps computed in sequence. 
At step $\ell \in \{ 1, \dots, m \}$ in the sequence, each graph edge feature $\mathbf{e}_{(i,j)}^{n, \ell}$ is updated using its value at the previous message-passing step $\ell-1$ and the values of the adjacent node features at step $\ell-1$, as follows
\begin{equation}
\label{processor1}
\mathbf{e}_{(i,j)}^{n, \ell}=\operatorname{mlp}_{e}^{\ell}\left([\mathbf{e}_{(i,j)}^{n,\ell-1}, \mathbf{v}_{i}^{n, \ell-1}, \mathbf{v}_{j}^{n, \ell-1}]\right), \qquad 
\ell \in \{ 1, \dots, m \}
\end{equation}
to obtain the updated value. 
In Eq.~(\ref{processor1}), the operator $\operatorname{[\cdot]}$ concatenates the given arguments on the feature dimension. 
The mapping in each step in the message passing is computed using MLP with residual connection and Rectified Linear Unit (ReLu) as the non-linear activation function.
Then, each graph node $\mathbf{v}_{i}^{n, \ell}$ is updated using its value at the previous message-passing step, $\ell-1$, and the aggregation of its incident edge features at step $\ell$:
\begin{equation}
\label{processor2}
\quad \mathbf{v}_{i}^{n, \ell}=\operatorname{mlp}_{v}^{\ell} \left( \operatorname{[} \mathbf{v}_{i}^{n,\ell-1}, \sum_{j \in \text{adj}(i)} \mathbf{e}_{(i,j)}^{n,\ell} \operatorname{]} \right), \qquad \ell \in \{ 1, \dots, m \}
\end{equation}
where $\text{adj}(i)$ is the set of nodes connected to node $i$.

Using Eqs.~(\ref{processor1})-(\ref{processor2}), the processor computes an updated set of node features that are then used by the Graph ConvLSTM to produce recurrent memories.  
The update of the edge-based messages, $\mathbf{e}_{(i,j)}^{n, \ell}$, is key to the accuracy of the MGN flow predictions as it propagates information between neighboring graph nodes (i.e., between neighboring mesh cells). 
%The processor computes a new set of node representations for the graph, which will be fed into the recurrent model, described below, for gaining the recurrent memories. Moreover, computing the edge-based messages, $\mathbf{e}_{i j}^{t, m}$, collects the rich interaction information between neighboring cells, which is demonstrated to benefit MGN's capability for predicting flow behaviors (). 
This design choice differentiates MGN from other classical GNN frameworks relying only on node features (see \cite{ZHOU202057}), such as GCN and GraphSAGE. 
Moreover, leveraging edge information makes it possible to capture nontrivial topological information regarding connectivity and transmissibility, which play an important role in HF simulations and could be used to infuse more physics into the data-driven model (see Section~\ref{sec:inject_physics}). 
\subsubsection{Graph-based convolutional LSTM model}
\label{subsec:graph_convLSTM}

%We use a recurrent GNN (RNN) model to capture temporal dynamics from the latent graph. Specifically, the RNN takes as input the node embeddings of a latent graph $\mathcal{G}^{t, m}$, and captures temporal dynamics by outputting recurrent memories extracted from previous states. 

To limit the temporal error accumulation and improve prediction accuracy, we complement MGN with a variant of Convolutional Long Short-Term Memory (ConvLSTM) that operates on graph data named Graph Convolutional LSTM (GConvLSTM) \cite{seo2018structured}. 
The latter is obtained by replacing the convolutional operator in ConvLSTM with a graph operator. 
Specifically, we follow the choice made in \cite{seo2018structured} and replace the Euclidean 2D convolution kernel with the Chebyshev spectral convolutional kernel \cite{defferrard2016convolutional}, whose hyperparameters are given in Appendix~\ref{app:lstm}. 
Graph spectral filters are known to perform effectively on graph-based data with a small number of parameters thanks to their isotropic nature \cite{seo2018structured}.
%
%\todo[inline]{MA: need to fill in here the appendix number. Also, why use spectral kernel?}
%
The goal of the Graph ConvLSTM step at time $t_n$ is to compute the cell state $\mathbf{C}^n$ and the hidden state $\mathbf{H}^n$. 
This is done as a function of the latent representation of node features computed  by the processor of MGN at time $t_n$ (Section \ref{subsec:encoder_processor}) and of the recurrent memories $\mathbf{C}^{n-1}$ and $\mathbf{H}^{n-1}$.
%
%This is done by applying sequences $\mathcal{F}$ and $\mathcal{G}$ of input gates, output gates, forget gates, and graph convolution operator as detailed in \cite{seo2018structured}:
%\begin{equation}
%\begin{aligned}
%\mathbf{C}^n &= g ( \boldsymbol{V}^{n,m}, \mathbf{C}^{n-1}, \mathbf{H}^{n-1} ), \\
%\mathbf{H}^n &= h ( \boldsymbol{V}^{n,m}, \mathbf{C}^{n}, \mathbf{H}^{n-1}),
%\end{aligned}
%\label{lstm}
%\end{equation}
%where $\boldsymbol{V}^{n,m} = [(\mathbf{v}^{n,m}_1)^T, \dots, (\mathbf{v}^{n,m}_{n_C})^T]^T$.
%
%After evaluating Eq.~(\ref{eq:lstm}), the updated $\mathbf{H}^{n}$ is decoded into the next-step physical state as explained in Section~\ref{subsec:decoder}.

Using the terminology of ConvLSTM, the GConvLSTM architecture involves a set of memory cells, namely the cell state $\mathbf{C}^n$ and the hidden state $\mathbf{H}^n$. 
GConvLSTM also relies on input gates $\mathbf{i}^n$, output gates $\mathbf{o}^n$, and forget gates $\mathbf{f}^n$ defined in \cite{seo2018structured}. 
These gates are based on a graph convolution operator and are used to control the flow of information into, out of, and within the memory cells. 
By construction, the cell state $\mathbf{C}^n \in \mathbb{R}^{n_C \times n_H}$ and the hidden state $\mathbf{H}^n \in \mathbb{R}^{n_C \times n_H}$ exhibit temporal dynamics and can contain spatial structural information of the graph-based input $\boldsymbol{V}^{n,m} = [(\mathbf{v}^{n,m}_1)^T, \dots, (\mathbf{v}^{n,m}_{n_C})^T]^T$ at time $t_n$. 
The functions of GConvLSTM are computed as follows: 

\begin{equation}
\begin{aligned}
\mathbf{i}^n &=\sigma\left( \mathbf{W}_{x i} *_{\mathcal{G}} \boldsymbol{V}^{n,m}+\mathbf{W}_{h i} *_{\mathcal{G}} \mathbf{H}^{n-1} + \mathbf{b}_i\right), \\
\mathbf{f}^n &=\sigma\left(\mathbf{W}_{x f} *_{\mathcal{G}} \boldsymbol{V}^{n,m}+\mathbf{W}_{h f} *_{\mathcal{G}} \mathbf{H}^{n-1} + \mathbf{b}_f\right), \\
\mathbf{C}^n &= \mathbf{f}^n \odot \mathbf{C}^{n-1} + \mathbf{i}^n \odot \tanh \left( \mathbf{W}_{x c} *_{\mathcal{G}} \boldsymbol{V}^{n,m} +\mathbf{W}_{h c} *_{\mathcal{G}} \mathbf{H}^{n-1}+\mathbf{b}_c\right), \\
\mathbf{o}^n &=\sigma\left(\mathbf{W}_{x o} *_{\mathcal{G}} \boldsymbol{V}^{n,m} +\mathbf{W}_{h o} *_{\mathcal{G}} \mathbf{H}^{n-1} +\mathbf{W}_{c o} \mathbf{C}^n + \mathbf{b}_o\right), \\
\mathbf{H}^n &= \mathbf{o}^n \odot \tanh \left(\mathbf{C}^n\right),
\end{aligned}
\label{eq:lstm}
\end{equation}
where $*_{\mathcal{G}}$ denotes the graph convolution operator, $\odot$ denotes the Hadamard product, and $\sigma(\cdot)$ is the sigmoid activation function. 
$\mathbf{W}$ and $\mathbf{b}$ are respectively the weights of the graph convolutional kernel and the bias term. %For instance,  $\mathbf{W}_{x i} *_{\mathcal{G}} \boldsymbol{V}^{n,m}$ denotes a graph convolution of $\boldsymbol{V}^{n,m}$ with $\mathbf{W}_{x i}$.
$\mathbf{i}^n$ determines how much of the new input is incorporated into the cell state. $\mathbf{f}^n$ controls the information to eliminate from the previous cell state. $\mathbf{o}^n$ determines how much of the cell state is output to the next time step.

In MGN-LSTM, the input vector $\boldsymbol{V}^{n,m} \in \mathbb{R}^{n_C \times n_H} $ is the latent representation of node features computed by the processor of MGN at time $t_n$ (Section \ref{subsec:encoder_processor}). 
In Eq.~(\ref{eq:lstm}), the vectors $\mathbf{C}^{n-1}$ and $\mathbf{H}^{n-1}$ are the recurrent memories obtained from the previous time step. 
After evaluating Eq.~(\ref{eq:lstm}), the updated $\mathbf{H}^{n}$ is decoded into the next-step physical state as explained in Section~\ref{subsec:decoder}. 
More details regarding GConvLSTM gates and states can be found in \cite{seo2018structured}.

\subsubsection{Decoder}
\label{subsec:decoder}
The decoder maps the updated hidden state $\mathbf{H}^{n}$ computed by GConvLSTM to the dynamic node-based properties in physical state using MLP as follows:
\begin{equation}
\label{decoder}
\hat{y}_{i}^{n+1}=\operatorname{mlp}_{v}^{m+1}\left(\mathbf{h}_{i}^{n}\right)
\end{equation}
where $\mathbf{h}_{i}^{n}$ contains the rows of the hidden state vector $\mathbf{H}^{n}$ corresponding to the updated latent vector of graph node $i$, and $m$ is the number of steps performed by the processor. 
Detailed illustrations of the encoder, processor, LSTM cell, and decoder are given in Appendices \ref{app:encoder}, \ref{app:processor}, and \ref{app:lstm}, respectively. 
% put an algorithm graph in the end to summarize the process

\subsection{Loss function and Optimizer}
\label{subsec:loss_function}
%\subsubsection{Loss function design}

We train MGN-LSTM on training time steps by minimizing the misfit between the true node label (HF simulation results) and the predicted node label. 
We use the per-node root mean square error (RMSE) loss to quantify the data mismatch for each time step.
The loss function reads: 
\begin{equation}
\mathcal{L}_{\text {MGN-LSTM}} =  \sum_{n=1}^{n_T} \sqrt{ \frac{1}{n_{B}} \sum_{i=1}^{n_{B}} \left\|y^{n}_i-\hat{y}^{n}_i\right\|_2^2}
\label{loss_fn}
\end{equation}
where $n_{B}$ denotes the number of nodes in a batch of training meshes, $y^{n}_i$ denotes the true output in the data set, $\hat{y}^{n}_i$ is the output predicted by MGN-LSTM, as formalized in Eq.~(\ref{eq-RMGN}). 

During training, the learning weights of $f_{\text{MGN-LSTM}, \theta}$ are updated based on the gradient to the loss function through back-propagation. 
Unlike next-step models such as MeshGraphNet~\cite{pfaff2020learning}, MGN-LSTM can propagate gradients throughout the entire sequence, allowing the model to utilize information from previous steps to produce stable predictions with small error accumulation. 
Furthermore, recurrent memories output by LSTM in MGN-LSTM retain information from previous inputs, which can be utilized to inform future predictions. 

An adaptive moment estimation (ADAM) optimizer is used and the learning rate is gradually decreased from $10^{-4}$ to $10^{-6}$.
With the dataset and a fixed set hyperparameters at hand, each epoch takes 40s to 100s on an NVIDIA A100 GPU depending on the architecture dimension. The value of specific training hyperparameters is given in Appendix~\ref{app:training_hyper}.
%
%
%In this study, we use `detrending' scaling \cite{tang2022deep} for all fields of node/edge features.  
%
%Specifically, the preprocessing can be expressed as: 

%
%  \begin{equation}
%     \tilde{\mathbf{x}}^{n}_i = \frac{\mathbf{x}^{n}_i -  \text{mean}([\mathbf{x}_1^n, \ldots,\mathbf{x}_{n_{S}}^n ])}{\text{std}([\mathbf{x}_1^n, \ldots,\mathbf{x}_{n_{S}}^n ]}, \quad i=1,\ldots, n_{S}, \ \ n = 1, \ldots, n_T,
%     \label{eq:p-0}
%\end{equation}
%
%where $n_{S}$ is the number of training samples. With this approach, a field of node features at time $t_n$, $\mathbf{x}^{n}_i$, can be normalized by subtracting the mean field (over all $n_{S}$ samples) and dividing each element by its standard deviation at that time step.

% Surrogate model evaluation

\section{Results}
\label{sec:results}

In this section, we first describe the dataset considered to train and test the MGN-LSTM model.
We illustrate that MGN-LSTM can accurately predict the CO$_2$ saturation plume and pore pressure evolution in the presence of impermeable faults.
At the end of this section, we also show how the saturation prediction accuracy can be improved by incorporating more physical properties (for instance, relative permeability) in the node and edge features of the graph.
The results of MGN-LSTM will be compared with those obtained with standard LSTM in Section~\ref{sec:comparison_with_standard_mgn}.

\subsection{Data description and MGN-LSTM training setup}
\label{sect:setup}

\begin{figure}[htbp!]
  \centering
  \includegraphics[trim={0cm 0cm 0cm 0cm},clip, scale=.80]{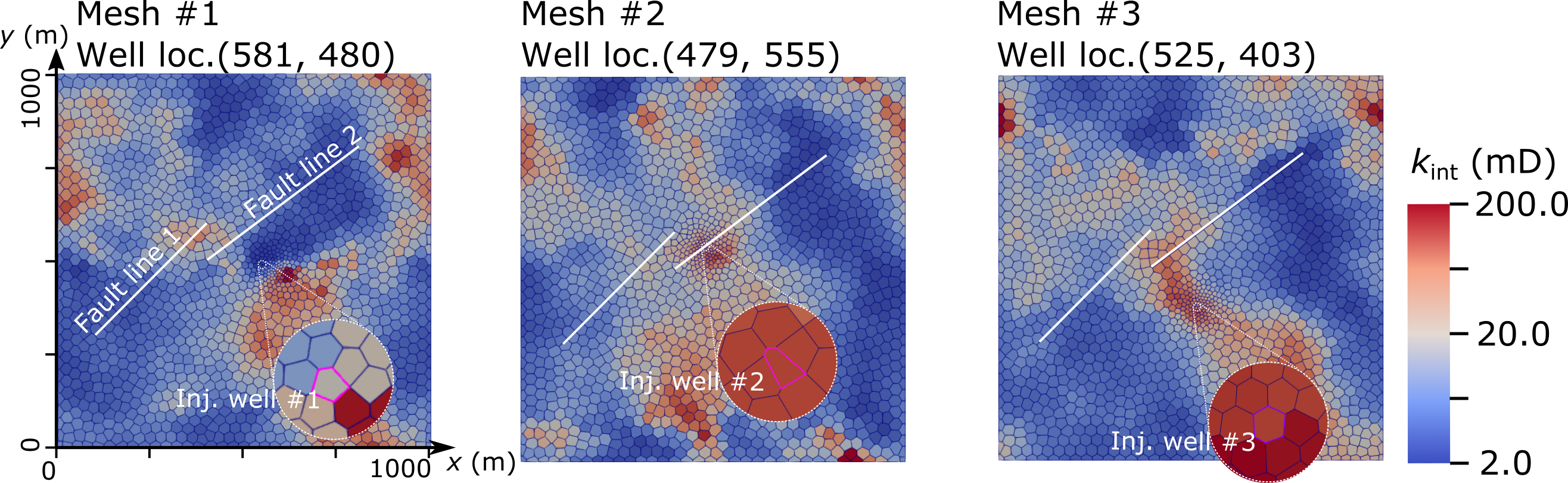}
   \caption{Heterogeneous permeability realizations with two fixed impermeable faults and one injection well for three cases. The coordinates of well location in each sampled case is shown on the top, and the insets show the enlarged vicinity of each well.} 
  \label{fig:data_setup}
\end{figure}

We generate a total of 500 realizations of the synthetic geological models of size 1 km x 1 km x 1 m.
The domain shape as well as the position of the two impermeable faults are fixed across all realizations. 
The coordinates of the endpoints of fault line 1 are (100m, 300m) and (400m, 600m), and the coordinates of the endpoints of fault line 2 are (400m, 500m) and (800m, 800m).
The synthetic models differ in their geological parameters (permeability), mesh configuration, and well location.
In each synthetic model, we first randomly generate the location of the injection well constrained within a prescribed 200 x 200 m box in the center of the domain to ensure that the injector is not placed too close to the boundary.
Then, a PEBI mesh conforming exactly to the specified faults and refined around the well is generated.
After that, we create a geomodel in which the permeability values are assigned to each cell according to a randomly generated Von Karman distribution~\cite{vonk2d} using SGeMS~\cite{sgems}. Specifically, the mean and standard deviation of log-permeability are 3.912 ln(mD) and 0.5 ln(mD), respectively, which results in an average permeability of 50 mD in the reservoir. A constant porosity of 0.2 is assigned to all cells. 
Fig.~\ref{fig:data_setup} shows three sampled geomodels and meshes used for training.
High-fidelity numerical simulation is then performed for each model using the GEOS simulator.

%\begin{table}
%\caption{Details of dataset for CO$_2$ storage operation}
%\centering
%\begin{tabular}{@{}ll@{}}
%\toprule
%\textbf{Dataset attributes}                                                & \textbf{Value} \\ \midrule
%Solver                                                                     & GEOS        \\
%Mesh type                                                                  & Unstructured PEBI mesh
%\\
%Number of time steps                                                       & 20           \\
%Time step                                                                  & 50 days
%\\
%Number of training trajectory                                              & 450
%\\
%\begin{tabular}[c]{@{}l@{}}Number of test trajectory\end{tabular}       & 50            \\ \midrule
%\textbf{Dataset statistics}                                                & \textbf{Value} \\ \midrule
%Number of graph nodes (avg.)                                                     & 1885           \\
%Number of graph node attributes                                                  & 8 
%\\
%Number of graph edges (avg.)                                                     & 4500
%\\
%Number of graph edge attributes                                                  & 3
%\\ \bottomrule
%\end{tabular}
%\label{table:dataset}
%\end{table}

%
MGN-LSTM is trained with 450 input meshes, in which each mesh has an 11-step rollout of simulation data, representing 550 days of CO$_2$ injection. 
The rollout division is linear with 50 days per step.
The trained model is then tested on 50 unseen meshes for 11-step rollouts and 19-step rollouts with temporal extrapolation to 950 days.  
The average number of graph nodes and graph edges in the dataset are 1885 and 4500 respectively.
%
%Separate models with the same architecture design are trained for predicting different dynamical quantities, which results in two MGN-LSTM models dedicated to predicting gas saturation and pore pressure, respectively.
We use two separate MGN-LSTM instances of the same architecture to predict the two different dynamical quantities, namely gas saturation and pore pressure.
The only difference between these two models is the dynamical quantity used to form node features and node labels during the training and inference stages. 
%
%The details of the dataset are provided in Table~\ref{table:dataset}.

In this study, we use `detrending' scaling \cite{tang2022deep} for all fields of node/edge features.  
Specifically, the preprocessing can be expressed as: 

  \begin{equation}
     \tilde{\mathbf{x}}^{n}_i = \frac{\mathbf{x}^{n}_i -  \text{mean}([\mathbf{x}_1^n, \ldots,\mathbf{x}_{n_{S}}^n ])}{\text{std}([\mathbf{x}_1^n, \ldots,\mathbf{x}_{n_{S}}^n ])}, \quad i=1,\ldots, n_{S}, \ \ n = 1, \ldots, n_T,
     \label{eq:p-0}
\end{equation}
where $n_{S}$ is the number of training samples. With this approach, a field of node features at time $t_n$, $\mathbf{x}^{n}_i$, can be normalized by subtracting the mean field (over all $n_{S}$ samples) and dividing each element by its standard deviation at that time step.

\subsection{Evaluation metrics}
\label{subsec:metric}

To quantify the prediction accuracy for gas saturation, we use the plume saturation error, $\delta^{s_{g}}$, introduced in~\cite{wen2022accelerating} and defined as: 

\begin{equation}
\begin{aligned}
& \delta^{s_{g}}=\frac{1}{\sum_{i,n} I^{n}_{i}} \sum_{n = 1}^{n_T} \sum_{i = 1}^{n_C} I^{n}_{i}\left|s^{n}_{g, i} - \hat{s}_{g, i}^n\right|, \\
& I_{i}^n =1 \quad \text { if } \quad\left(s_{g,i}^{n}>0.01\right) \cup\left(\left|\hat{s}^n_{g,i}\right|>0.01\right),
\end{aligned}
\label{gs_error}
\end{equation}
where $I^{n}_{i}=1$ indicates that a mesh cell has a non-zero gas saturation in either the ground truth or the prediction, $s^n_g$ denotes the gas saturation from HF simulations (ground truth), $\hat{s}^n_g$ is the predicted gas saturation, $n_T$ is the number of temporal snapshots, including training ranges (11 steps, 550 days) and extrapolated ranges (8 steps, 400 days), and $n_C$ is the number of cells in the mesh, which can vary between simulation models.
As discussed in~\cite{wen2022accelerating}, $\delta^{s_g}$ provides a strict metric to evaluate CO$_2$ gas saturation because it focuses on the accuracy within the separate phase plume. 

We use the relative error $\delta^{p_g}$ defined below to evaluate the prediction accuracy for pore pressure:
\begin{equation}
\begin{aligned}
\delta^{p_g}=\frac{1}{n_{C} n_T} \sum_{n = 1}^{n_T} \sum_{i = 1}^{n_C} \frac{\left|p_{g,i}^{n} -\hat{p}_{g,i}^n\right|}{p_{ g,\textit{init}}},
\end{aligned}
\label{pp_error}
\end{equation}
where $p^n_g$ denotes the ground truth pore pressure given by the HF simulation, $\hat{p}^n_g$ is the predicted pore pressure, and $p_{ g,\textit{init} }$ is the initial reservoir pressure, which remains identical ($10$ MPa) for all test/training cases.

In the following sections, we use these metrics to illustrate the accuracy of MGN-LSTM in two steps. 
First, in Section~\ref{subsec:prediction}, we consider a representative mesh in the test set and demonstrate the ability of MGN-LSTM to capture the complex plume dynamics for time steps beyond the training period.
Then, in Section~\ref{subsec:unseen_meshes}, we consider 10 representative test meshes to illustrate the reliability of MGN-LSTM predictions for unseen well locations, permeability fields, and meshes yielding very different CO$_2$ plume shapes.

\subsection{Predicting complex spatio-temporal dynamics beyond the training period}
\label{subsec:prediction}
\begin{figure}[htbp!]
  \centering
  \includegraphics[trim={0cm 0cm 0cm 0cm},clip, scale=.80]{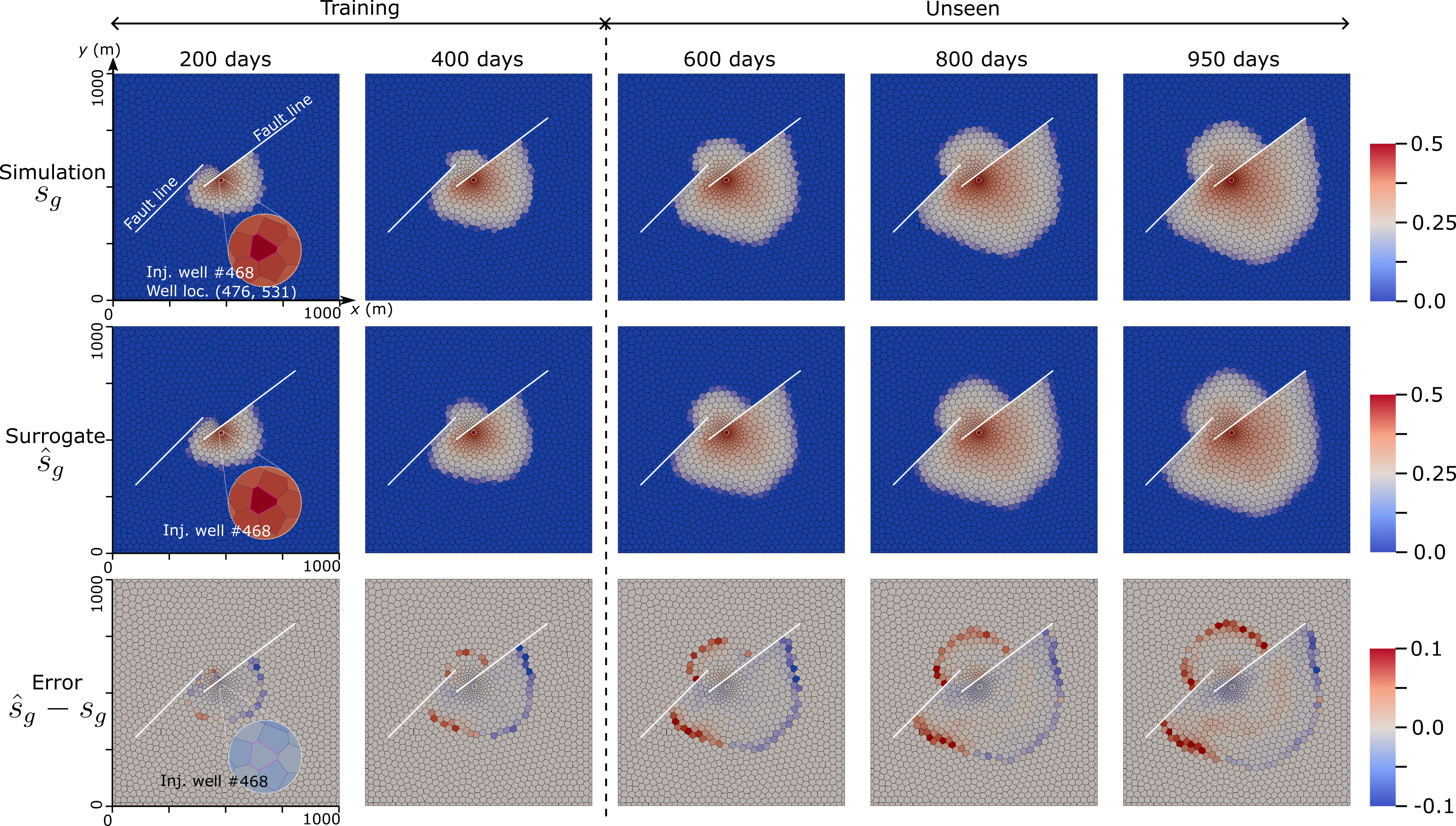}
   \caption{Temporal evolution of the CO$_2$ saturation plume. The first and second rows respectively show the CO$_2$ saturation fields from HF and MGN-LSTM for a representative test case (mesh 468) at five different times. The third row shows the saturation error between MGN-LSTM and HF. The first three columns represent predictions within the training period and the last two columns represent the model predictions for time steps beyond the training period.} 
  \label{fig:gnn_sg_mesh468}
\end{figure}

\begin{figure}[htbp]
  \centering
  \includegraphics[trim={0cm 0cm 0cm 0cm},clip, scale=1.0]{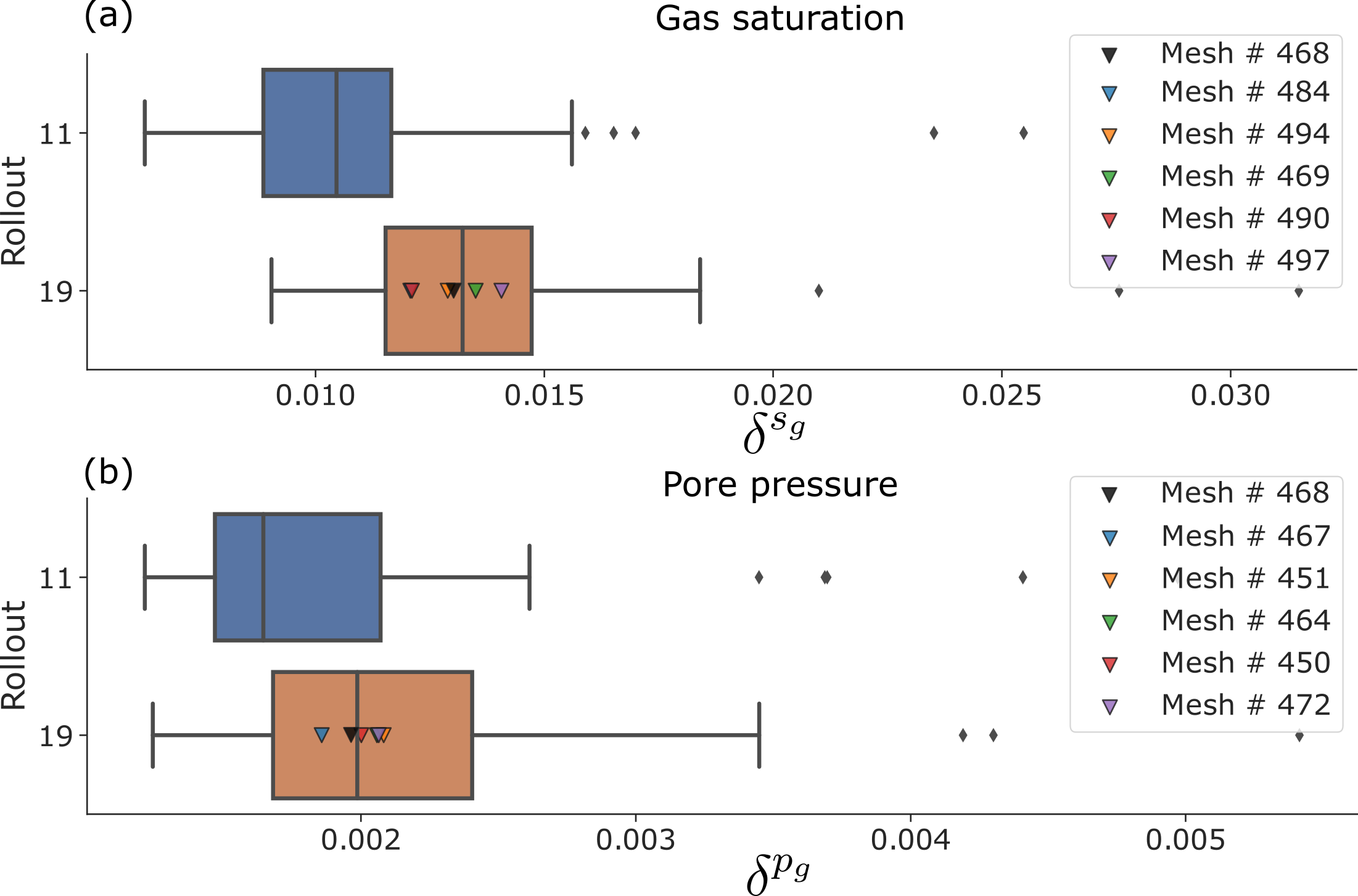}
   \caption{Accuracy of the MGN-LSTM prediction of (a) gas saturation and (b) pore pressure after rolling out 550 and 950 days for all test meshes. The meshes discussed in Section~\ref{subsec:unseen_meshes} are highlighted as triangles and shown on top of the boxplot of the 950-day rollout results.} 
  \label{fig:gnn_mesh_boxplot}
\end{figure}

In this section, we consider mesh 468 from the test set as an example.
The mesh and location of the well are highlighted in the insets of Fig.~\ref{fig:gnn_sg_mesh468}.
This test case has a CO$_2$ plume saturation error (Eq.~(\ref{gs_error})) within the interquartile range of the test ensemble and can therefore be considered as representative of the accuracy of the surrogate model predictions (see Fig.~\ref{fig:gnn_mesh_boxplot}). 
This particular configuration is illustrated here because its well location is close to the faults and therefore produces an interesting saturation plume.

\begin{figure}[htbp]
  \centering
  \includegraphics[trim={0cm 0cm 0cm 0cm},clip, scale=.80]{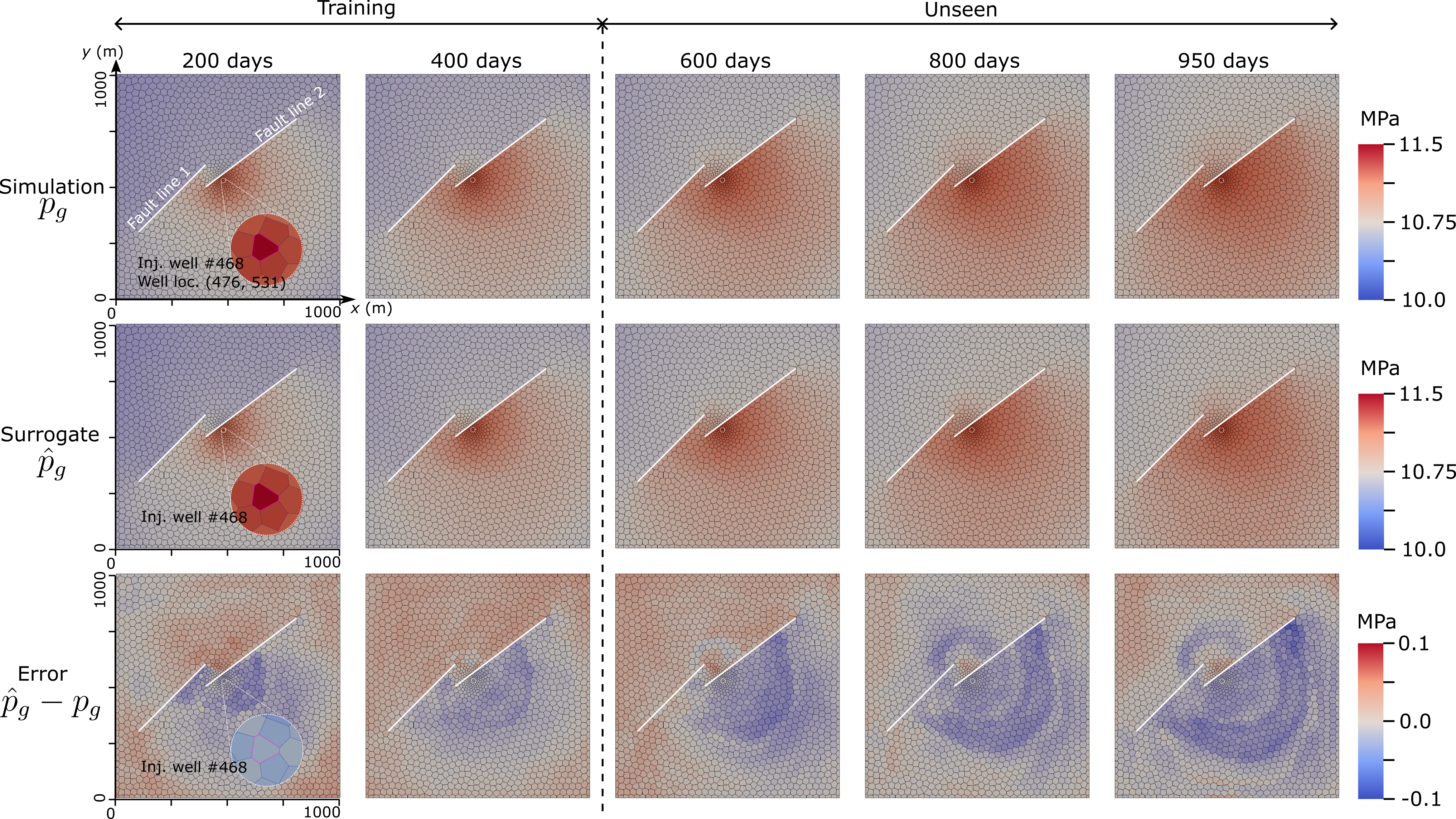}
   \caption{Temporal evolution of the pore pressure. The first and second rows respectively show the pore pressure fields from HF and MGN-LSTM for a representative test case (mesh 468) at five different times. The third row shows the pressure error between MGN-LSTM and HF. The first two columns show predictions within the training period and the last three columns show the model predictions for time steps beyond the training period.} 
  \label{fig:gnn_pore_pressure_mesh468}
\end{figure}

Figure~\ref{fig:gnn_sg_mesh468} compares the MGN-LSTM prediction of the CO$_2$ saturation fields with the HF simulation results at five snapshots ($t$ = 200, 400 days in the training period, $t$ = 600, 800, 950 days beyond the training period). 
MGN-LSTM and HF simulations exhibit an excellent match.
Despite the presence of impermeable faults near the injector, we observe that (1) MGN-LSTM can accurately capture the complex temporal evolution of the CO$_2$ plume in the presence of impermeable faults, and (2) MGN-LSTM can extrapolate beyond the training horizon with a mild error accumulation at the saturation front.
The MGN-LSTM pore pressure predictions are presented in Fig~\ref{fig:gnn_pore_pressure_mesh468}.
They also exhibit a very good agreement with the HF simulation results during and after the training period.
Unlike for the saturation variable, the prediction errors are distributed over the entire domain due to the elliptic behavior of the pressure variable.

\subsection{Generalizability to meshes, well locations, and permeabilities not seen during training}
\label{subsec:unseen_meshes}

In this section, we consider 10 representative meshes from the test set (meshes 484, 494, 469, 490, 497 for the saturation and meshes 467, 451, 464, 450, 472 for pressure) to demonstrate that MGN-LSTM generalizes well for different meshes, boundary conditions (well locations) and permeability fields that are not included in the training set. 
As shown in Fig.~\ref{fig:gnn_mesh_boxplot}, the 10 test cases have a saturation error within the interquartile range of the test ensemble and are therefore representative of the surrogate accuracy. 

\begin{figure}[htbp]
  \centering
  \includegraphics[trim={0cm 0cm 0cm 0cm},clip, scale=.80]{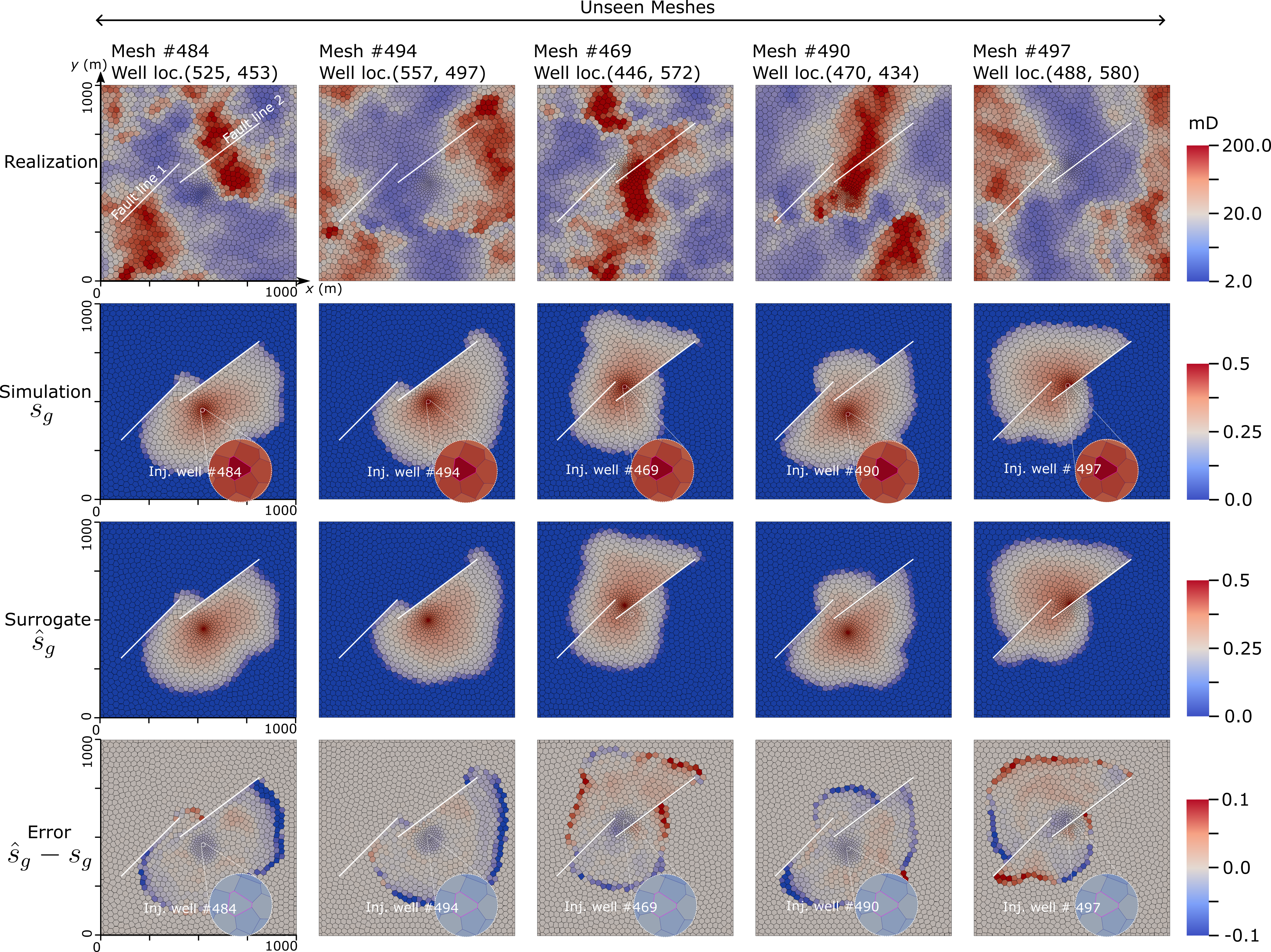}
   \caption{CO$_2$ saturation predictions at 950 days for five distinct test meshes. From top to bottom, the rows show reservoir permeability, HF simulation of CO$_2$ saturation, MGN-LSTM prediction of CO$_2$ saturation, and MGN-LSTM prediction errors. The coordinates of well location in each test case is shown on the top of first row and the insets show the enlarged vicinity of each well.} 
  \label{fig:gnn_inference_meshes_saturation}
\end{figure}

\begin{figure}[htbp]
  \centering
  \includegraphics[trim={0cm 0cm 0cm 0cm},clip, scale=.80]{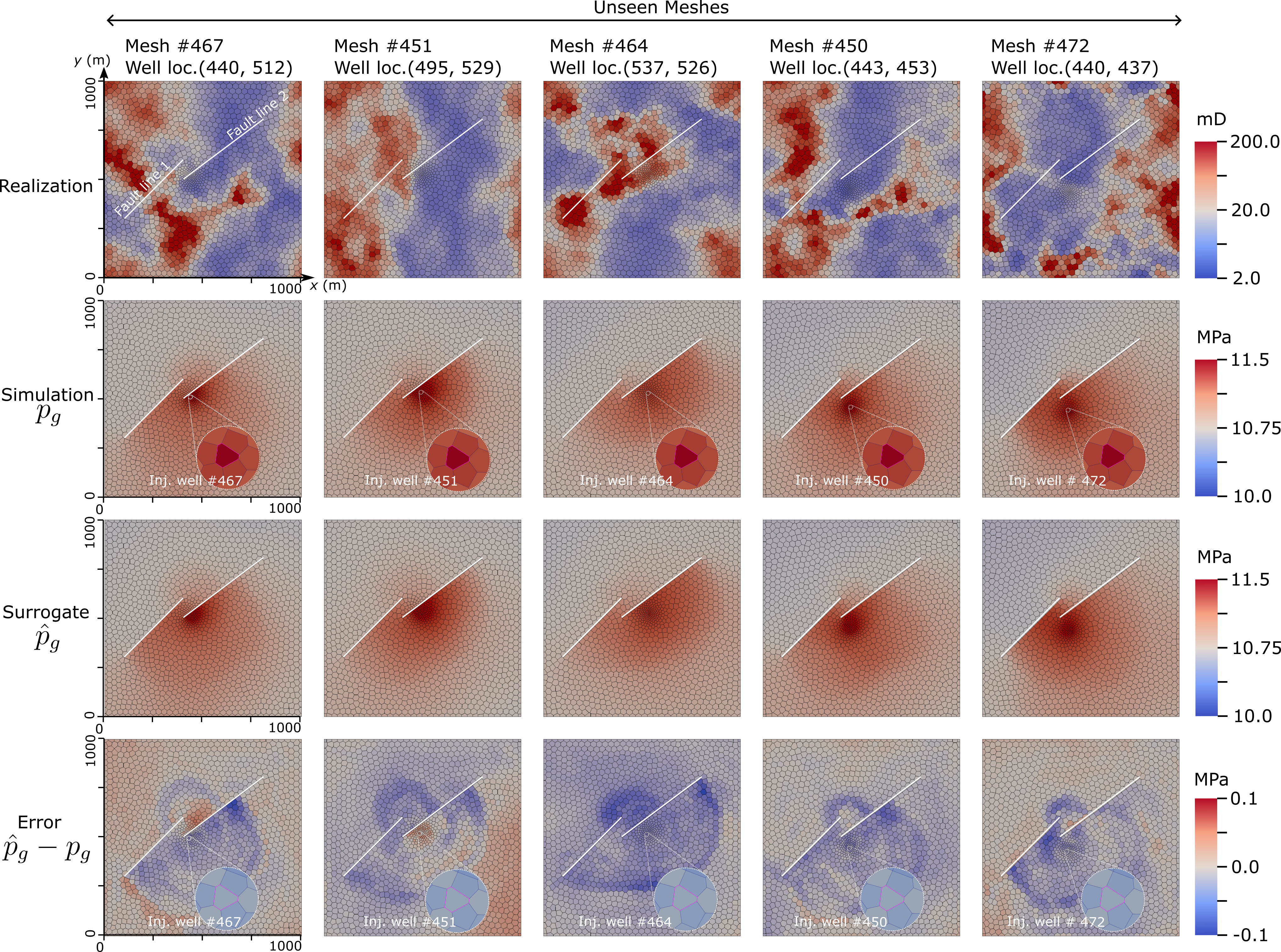}
   \caption{Pore pressure predictions at 950 days for five distinct test meshes. From top to bottom, the rows show reservoir permeability, HF simulation of pore pressure, MGN-LSTM prediction of pore pressure, and MGN-LSTM prediction errors. The coordinates of well location in each test case is shown on the top of first row and the insets show the enlarged vicinity of each well.} 
  \label{fig:gnn_inference_meshes_pressure}
\end{figure}

The results after 950 days are presented in Figs.~\ref{fig:gnn_inference_meshes_saturation}-\ref{fig:gnn_inference_meshes_pressure} for the CO$_2$ saturation and the pore pressure, respectively.
The figures show that the permeability field and well location relative to the faults vary drastically from one case to the other.
Due to the complex interplay between the saturation front and the faults, the differences in initial setup yield very different plume shapes.
Still, MGN-LSTM achieves an excellent agreement with the HF simulation results after 950 days for both saturation and pressure.
MGN-LSTM displays a remarkable generalizability considering the high dimensionality of the problem and the training data size. 
The median saturation errors in the CO$_{2}$ plume prediction for the training set and the extrapolated ranges in the testing set are only $1.0\%$ and $1.3\%$, respectively. 
This degree of accuracy in the prediction of CO$_2$ plume migration is sufficient for many practical applications, including the estimation of sweep efficiencies ~\cite{wen2022accelerating}.
Similarly, the pressure prediction exhibits an excellent accuracy. 
The median pore pressure errors for the training set and the extrapolated ranges in the testing set stand at $0.16\%$ and $0.19\%$.
This confirms the reliability of MGN-LSTM predictions for complex configurations unseen during training.

\subsection{Improving performance using augmented physics-based graph node/edge features}
\label{sec:inject_physics}

In this section, we describe a methodology to improve the prediction accuracy of MGN-LSTM.
In the previous sections, the edge feature of MGN-LSTM only accounted for distance-related information (see Section~\ref{subsec:graph_rep}). 
But, more physics insights can be infused into the algorithm by incorporating more information in the edge and node features.
Specifically, we explore three modifications of the gas saturation model discussed in earlier sections, focusing on the incorporation of additional features. 
The first modification involves the inclusion of static transmissibility ($T_{ij}$) as an additional edge feature, the second variation introduces phase relative permeability ($k_{r,i}$) as an additional node feature, and the third variation combines both static transmissibility and phase relative permeability as extra edge and node features, respectively. 
%
%For conciseness, these three modified cases are denoted as ``Trans.", ``Rel. Perm.", and ``Trans. $\&$ Rel. Perm." cases, respectively.
%
The transmissibility  remains fixed throughout rollout steps. 
%For the PEBI meshes used in all simulations, $A$ and $k_{ij}$ terms are the area of the common face and the harmonic weighted permeability between cell $i$ and $j$, respectively.
%$Note $T_{ij}$ is a static parameter throughout rollout steps. 
%
The relative permeability of gas phase $k_{r,i}$ is computed as a function of the predicted gas saturation, $s_{g,i}^{n-1}$. 
The input and output variables for each case are summarized in Table~\ref{tab:node_edge_feature}. 
%
%\todo[inline]{i think we need this piece of information. the question is where to put them. do we need to give this remark here or we put them in the appendix? }
%

\begin{table}[!hpb]
\caption{Input and output of the surrogate model: $s_{g,i}^n, V_i, k_i, \boldsymbol{n}_i, \boldsymbol{x}_i, k_{r,i}(s_{g,i}^{n-1})$ denote gas saturation, cell volume, scalar permeability, cell type, cell center, cell relative permeability, respectively, for a given cell. 
Note that $k_{r,i}$ takes the gas saturation at the previous prediction step, $s_{g,i}^{n-1}$, to update its value.
Here, $\boldsymbol{n}_i$ is a one-hot vector of size 4, encoding whether a cell is an internal cell, injector, cell along fault lines, and boundary cell.
$T_{i,j}$ denotes the static transmissibility between cells $i$ and $j$. 
}
\begin{center}
\begin{tabular}{|c|c|c|c|}
\hline Case & \makecell{Node \\ input} & \makecell{Edge \\ input} & \makecell{Node \\ output} \\
\hline Baseline & $s_{g,i}^n, V_i, k_i, \boldsymbol{n}_i, \boldsymbol{x}_i $ & \makecell{$\boldsymbol{x}_j-\boldsymbol{x}_i$, \\
$\left|\boldsymbol{x}_j-\boldsymbol{x}_i\right|$} & $s_{g,i}^{n+1}$ \\
\hline Static transmissibility & $s_{g,i}^n, V_i, k_i, \boldsymbol{n}_i, \boldsymbol{x}_i $ & \makecell{$\boldsymbol{x}_i-\boldsymbol{x}_j$, \\
$\left|\boldsymbol{x}_i-\boldsymbol{x}_j\right|$, \\
$T_{i,j}$} & $s_{g,i}^{n+1}$ \\
\hline Relative permeability & $s_{g,i}^n, V_i, k_i, \boldsymbol{n}_i, \boldsymbol{x}_i, k_{r,i}(s_{g,i}^{n-1})$ & \makecell{$\boldsymbol{x}_i-\boldsymbol{x}_j$, \\
$\left|\boldsymbol{x}_i-\boldsymbol{x}_j\right|$}  & $s_{g,i}^{n+1}$ \\
\hline \makecell{\text{Static transmissibility} \\
\text{Relative permeability}} & $s_{g,i}^n, V_i, k_i, \boldsymbol{n}_i, \boldsymbol{x}_i, k_{r,i}(s_{g,i}^{n-1})$ & \makecell{$\boldsymbol{x}_i-\boldsymbol{x}_j$, \\
$\left|\boldsymbol{x}_i-\boldsymbol{x}_j\right|$, \\
$T_{i,j}$ }  & $s_{g,i}^{n+1}$ \\ \hline
\end{tabular}
\end{center}
\label{tab:node_edge_feature}
\end{table}

\begin{figure}[htbp]
  \centering
  \includegraphics[trim={0cm 0cm 0cm 0cm},clip, scale=1.10]{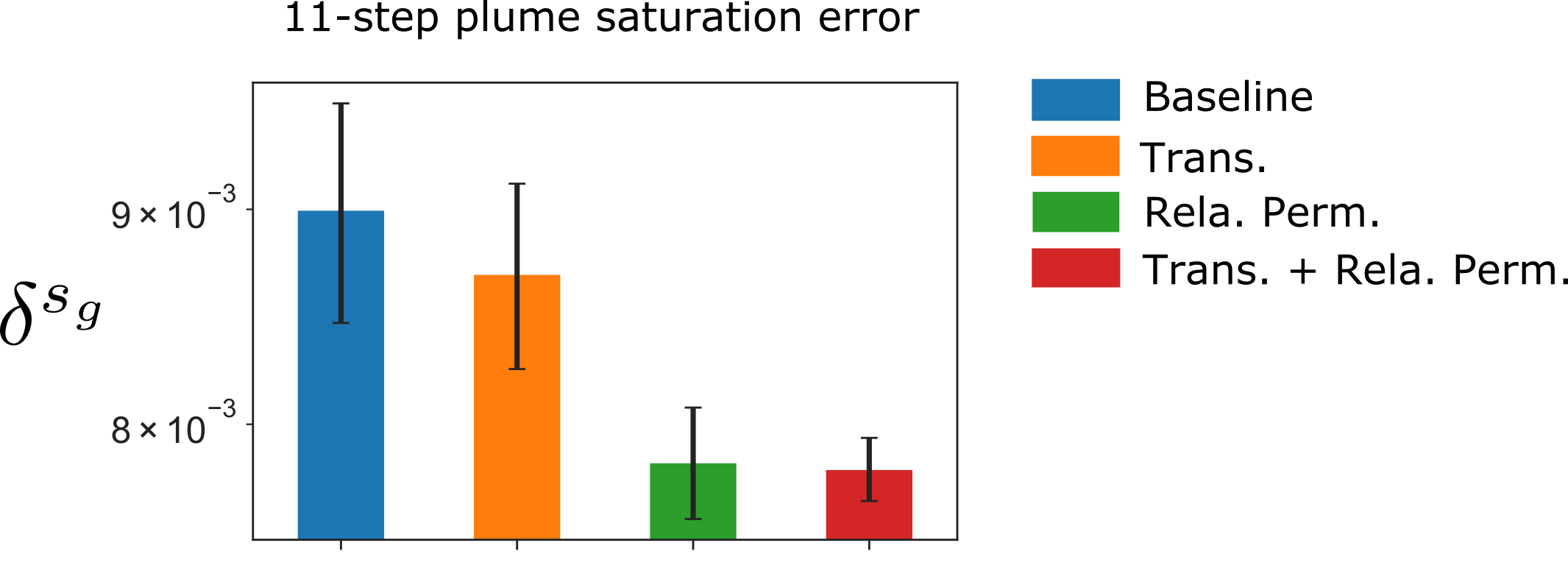}
  \caption{Effects of injecting more physics information on the 11-step rollout (550 days of injection) performances of MGN-LSTM at predicting CO$_2$ saturation. The error bars correspond to min/median/max performance across 12 random seeds.
  "Trans" denotes the case with transmissibility included in the edge features, "Rela. Perm." denotes the case with relative permeability included in the node features, and "Trans + Rela. Perm." combines the two approaches.} 
  \label{fig:inject_physics}
\end{figure}

These three modifications are evaluated separately and incrementally, and are only applied to the prediction of CO$_2$ saturation. 
We train each experiment with different random seeds. 
We evaluate the effect of augmented features on model performance by measuring the 11-step rollout plume error, which only includes the training range, namely the first 550 days.
Figure \ref{fig:inject_physics} confirms that adding more physical information in the node/edge features clearly improves CO$_2$ saturation prediction accuracy.
Using the transmissibility as an edge yields a mild error reduction in $\delta^{s_{g}}$, while the largest improvement is observed for the addition of relative permeability in the node feature with a near 10\%-reduction in $\delta^{s_{g}}$. This experiment demonstrated that incorporating more physics into the architecture can improve prediction accuracy even in a purely data-driven framework.

% Discussion
\section{Discussion}

\subsection{Comparison between MGN-LSTM and standard MGN}
\label{sec:comparison_with_standard_mgn}

The standard MGN approach \cite{pfaff2020learning} suffers from temporal error accumulation and often requires mitigation strategies to maintain a stable long-term prediction. 
In the present study, we rely on a noise injection technique~\cite{han2022predicting, pfaff2020learning} to enable a stable rollout of standard MGN.
We compare standard MGN and MGN-LSTM with the same encoder-processor-decoder architecture.
In MGN-LSTM, two separate standard MGN models are trained for predicting gas saturation and pore pressure. 
The accuracy improvements discussed in Section~\ref{sec:inject_physics} are not used here.
%
%\todo[inline]{MA: more appendix stuff that is missed}
The noise injection strategy and training procedure for MGN are detailed in Appendix~\ref{app:MGN}, as well as the model parameters used in the study.
The goal of this ablation-nature comparison is to demonstrate the ability of LSTM to constrain temporal errors.

%
%
%Two separate MGN models are trained for predicting gas saturation and pore pressure. The NI strategy and training procedure for MGN are detailed in appendix . The model parameters of all models used in this comparison are given in appendix .

%\subsubsection{Predicting spatiotemporal dynamics}
%\label{subsubsec:comparison_tempo}

%Figure 9 compares the performances of LSTM-MGN and MGN on mesh 468 for predicting gas saturation and pore pressure. Overal speaking, LSTM-MGN tends to predict both fields of dynamical quantities with temporal error well constrained. But, without the LSTM cell, MGN tends to make gas saturation predictions drastically deviating from the ground truth after 800 days of injection. This discrepancy is observed to be even more pronounced for predicting pore pressure, as the last two rows of Figure 9 demonstrate. MGN generates large prediction errors throughout the prediction, whereas LSTM-MGN predicts at high accuracy. 

%Error results, represented by the average value of , for both quantities are also shown in Fig. 10. It is evident that errors of gas saturation are generally lower, and that they grow less quickly in time, using our LSTM-MGN than with the baseline model. Analogous behavior is also observed for predicting pore pressure where errors with MGN are generally larger than with LSTM-MGN.

\label{subsec:gnn_tempo_boxplot}
\begin{figure}[htbp]
  \centering
  \includegraphics[trim={0cm 0cm 0cm 0cm},clip, scale=0.8]{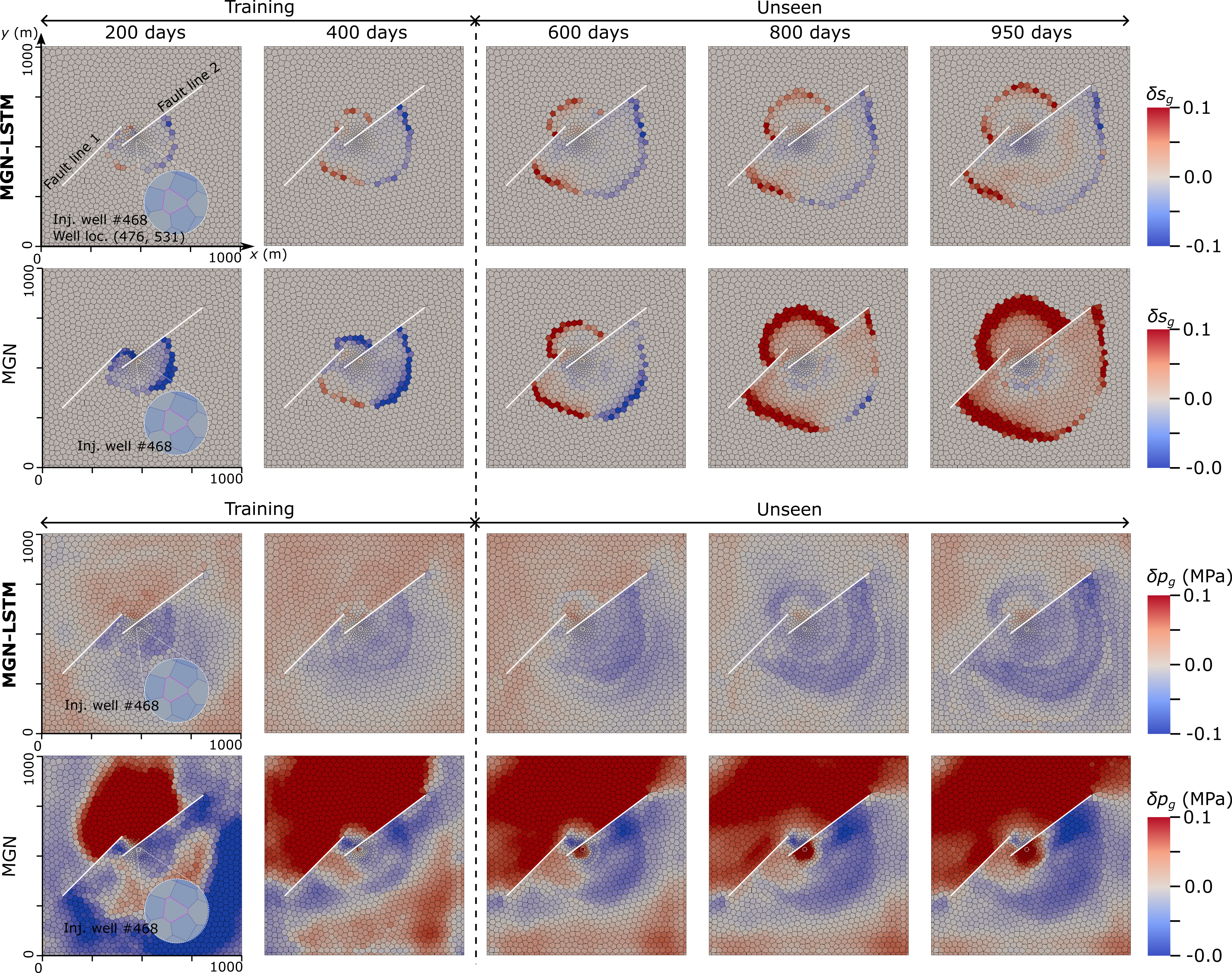}
   \caption{Prediction accuracy for (a) gas saturation and (b) pore pressure after rolling out 550 and 950 days for mesh 468 (see Section~\ref{subsec:prediction}).} 
  \label{fig:mgn_lstm_temp_comp}
\end{figure}

\begin{figure}[htbp]
  \centering
  \includegraphics[trim={0cm 0cm 0cm 0cm},clip, scale=0.8]{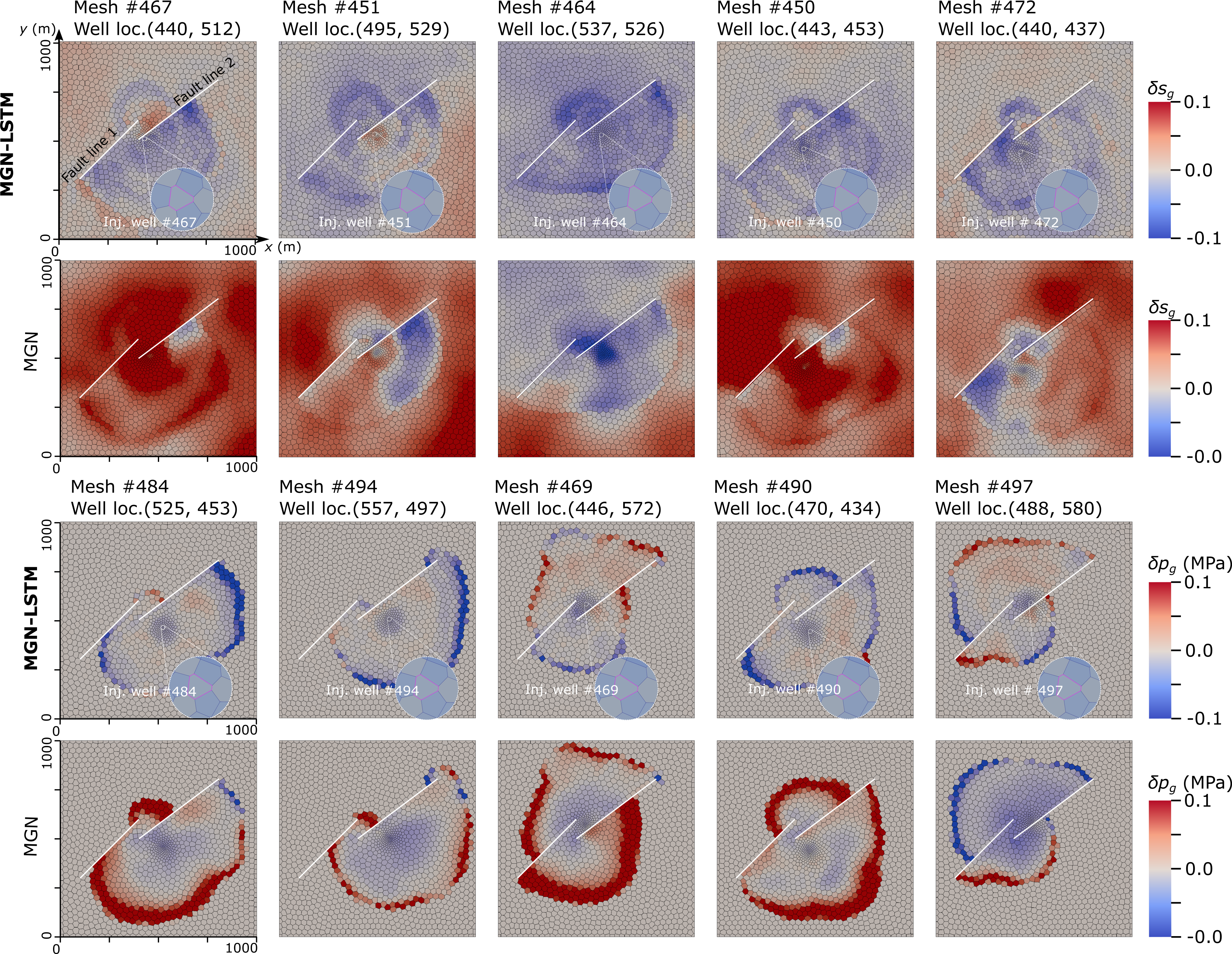}
   \caption{Prediction accuracy for (a) gas saturation and (b) pore pressure after rolling out 550 and 950 days for all test meshes of Section~\ref{subsec:unseen_meshes}.} 
  \label{fig:mgn_lstm_mesh_comp}
\end{figure}

Next, we briefly illustrate the difference in prediction accuracy between MGN-LSTM and standard MGN for the meshes considered in Sections \ref{subsec:prediction} and \ref{subsec:unseen_meshes}.
Considering mesh 468 of Section~\ref{subsec:prediction} first, we compare the predictions of standard MGN and MGN-LSTM after the training period in Fig.~\ref{fig:mgn_lstm_temp_comp}.
It is clear that for both saturation and pressure, the standard MGN predictions start deviating significantly from the HF data after 800 days.
The standard MGN error accumulates at the front for CO$_2$ saturation and is more diffused for pressure. 
In both cases however, the MGN-LSTM error remains constrained thanks to the addition of the LSTM cell.
Considering now the 10 meshes of Section~\ref{subsec:unseen_meshes}, Fig.~\ref{fig:mgn_lstm_mesh_comp} confirms the poor prediction accuracy of standard MGN after 950 days of injection for both CO$_2$ saturation and pressure compared to MGN-LSTM.

%\subsubsection{Generalizability to unseen simulation meshes, well locations, permeability heterogeneities}
%\label{subsubsec:comparison_tempo_unseen}
%The comparison between LSTM-MGN and MGN for predicting gas saturation and pore pressure on various meshes is shown in Figure 10. Note that in each testing mesh, the well location and mesh configuration is different. It is evident that MGN cannot predict pore pressures accurately whereas LSTM-MGN can generalize well to various well locations and unseen meshes. For predicting gas saturation, LSTM-MGN again generalizes well to unseen meshes and boundary conditions, whereas MGN tends to accumulate significant temporal errors around gas plume edges (see gas saturation predictions on Mesh 469 and 490). These empirical observations demonstrate the positive effect of LSTM cell on constraining temporal errors compared with one-step models, such as MGN.

\label{subsec:mgn_rmgn_mesh_comp}

\begin{figure}[htbp]
  \centering
  \includegraphics[trim={0cm 0cm 0cm 0cm},clip, scale=1.0]{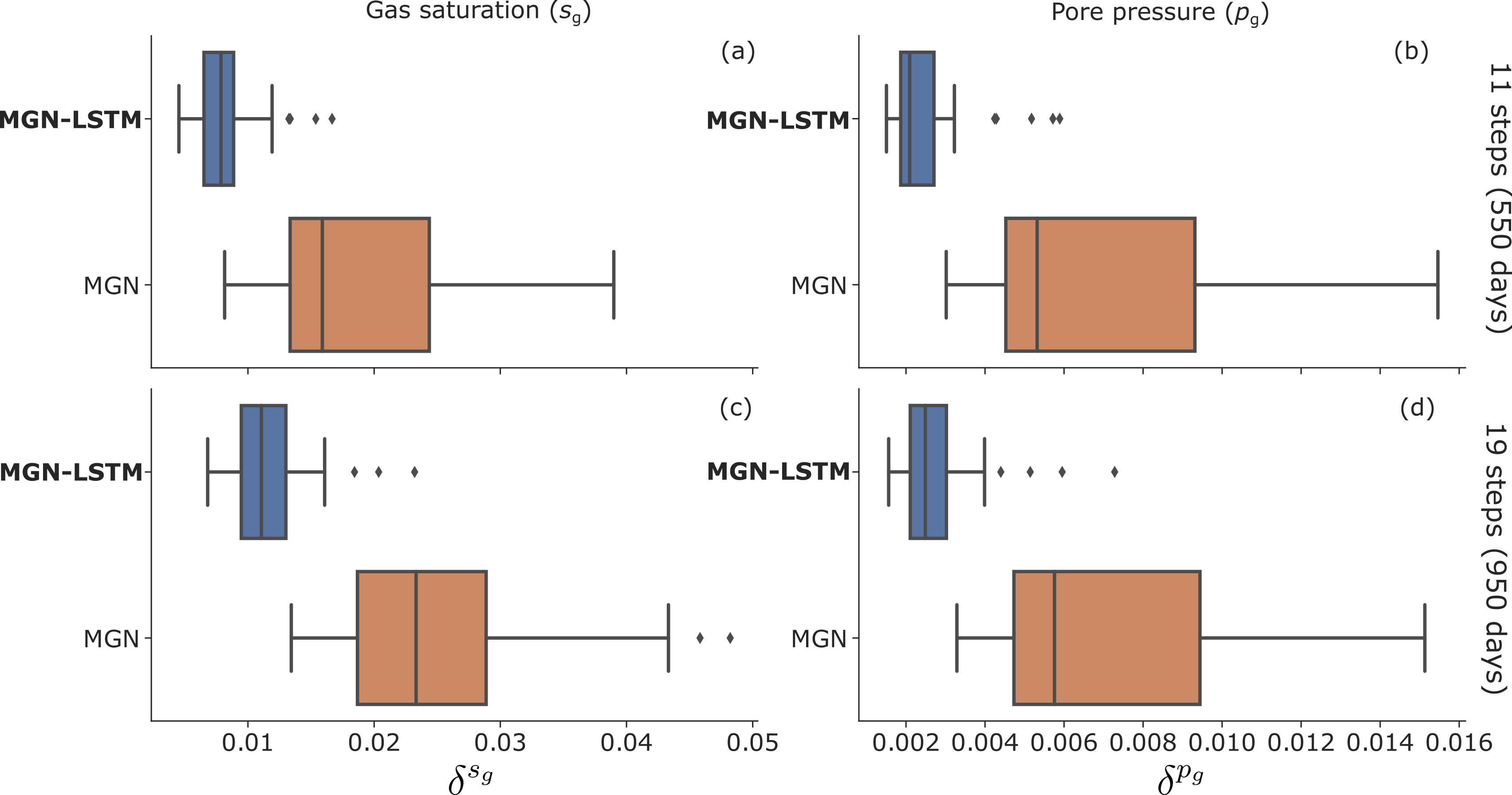}
   \caption{Boxplots of the test ensembles for plume saturation and pressure errors of MGN-LSTM and standard MGN. Panels (a) and (c) compare the CO$_2$ saturation predictions of MGN-LSTM and standard MGN after rolling out 11 steps (550 days) and 19 steps (950 days), respectively, on the test ensembles. Panel (b) and (d) compare the pore pressure predictions of MGN-LSTM and standard MGN after rolling out 11 steps (550 days) and 19 steps (950 days), respectively, on the test ensembles. MGN-LSTM is in orange and standard MGN is in blue.}
  \label{fig:gnn_rmse}
\end{figure}

%\subsubsection{Ensemble performances between RMGN and MGN}
We summarize the respective accuracy of MGN-LSTM and standard MGN by comparing their ensemble results. 
Figure~\ref{fig:gnn_rmse} shows the boxplots of error metrics $\delta^{S_g}$ and $\delta^{P_g}$ evaluated at (1)  550 days (the end of training) and (2) 950 days (400 days of extrapolation).
We observe that in both cases, MGN-LSTM outperforms standard MGN in terms of accuracy for the prediction of CO$_2$ saturation as well as pressure. 
This is particularly the case after the end of the training period (i.e., between 550 days and 950 days), as the standard MGN saturation prediction accuracy deteriorates significantly due to error accumulation.

%Figure\ref{fig:gnn_rmse} shows the comparison result, where the boxplots of RMSEs for two models after rolling out gas saturation/pore pressures for 550/950 days are shown (see Fig. 4(a) and (c)).  Under both scenarios, namely 11-step and 19-step rollouts, RMGN outperforms the strong baseline, MGN in terms of the RMSEs of saturation prediction. Especially when a longer rollout is involved, RMGN only shows a minor increase of temporal error whereas MGN experiences a large error accumulation. 

%A similar observation is found for the pressure prediction where RMGN models outperform the baseline model (see Fig. 4(b) and (d) and Table. 2). Although both models show to be capable of predicting pore pressure in future steps without further accumulating errors, under long-term rollouts RMGN still performs better than that of MGN. This experiment clearly demonstrates the graph-based recurrent component blended with MGN benefits constraining the temporal error accumulations suffered by ”one-step” GNN  models with noises being injected. 

\subsection{Computational efficiency}
To analyze the computational efficiency, we compare the inference times of MGN-LSTM, standard MGN, and the high-fidelity simulator, GEOS, as demonstrated in Table~\ref{tab:gnn_inference}. Utilizing the previously discussed dataset with 1,885 cells per mesh on average, MGN-LSTM requires an average of 0.31 seconds for a 19-step rollout on an NVIDIA Tesla A100 GPU to process a single batch. This contrasts with standard MGN's shorter average time requirement of 0.07 seconds on the same hardware. The inference time for MGN-LSTM is around four times that of standard MGN, primarily due to the auto-regressive prediction of dynamic predictions in LSTM. The reduction of this overhead is an area of focus for future work.
Comparing these surrogate models with GEOS demonstrates a significant performance gain. Specifically, on the same dataset, MGN-LSTM exhibits a nearly 160-fold reduction in execution time in comparison to GEOS, which operates on a CPU Intel(R) Xeon(R) E5-2680 v4 2.10GHz. We expect that this performance gain will be even more prominent with a larger mesh size.

\begin{table}[!hpt]
    \begin{threeparttable}
        \caption{Average inference times for MGN-LSTM and standard MGN with 11-step rollout (550 days) and 19-step rollout (950 days). The inference times are compared with the GEOS run time after 550 days and 950 days.}
        \begin{center}
        \begin{tabular}{|c|c|c|c|}
        \hline  & \makecell{MGN-LSTM avg.\\ inference time (s)\tnote{a}} & \makecell{Standard MGN avg.\\ inference time (s)\tnote{b}} & \makecell{GEOS \\ run time (s)\tnote{c}} \\
        \hline 11-step rollout (550 days)  & 0.18 & 0.04 & 22.12 \\
        \hline 19-step rollout (950 days) &  0.31 & 0.07 & 49.02 \\ \hline
        \end{tabular}
        \begin{tablenotes}
            \item[a] On an NVIDIA Tesla A100 GPU, single-batch inference run
            \item[b] On an NVIDIA Tesla A100 GPU, single-batch inference run
            \item[c] On an Intel Xeon E5–2695 v4, single-core serial run
        \end{tablenotes}
        \end{center}
        \label{tab:gnn_inference}
    \end{threeparttable}
\end{table}

\section{Concluding remarks}
\label{sec:conl}
This paper aims at demonstrating the applicability of deep-learning-based surrogate models for subsurface flow simulations with complex geological structures.
We present a graph-based neural surrogate model that can operate on unstructured meshes and naturally handle geological fault structures. 
Our model combines a graph-based Long-Short-Term-Memory (LSTM) cell with a one-step graph neural network model, MeshGraphNet (MGN), to control temporal error accumulation.
The model is trained on 450 high-fidelity, unstructured simulation results.
The mesh configuration, well location, and reservoir permeability field vary from one simulation to the other. 

The accuracy and performance of MGN-LSTM were analyzed on a set of 50 test cases.
Our results demonstrate that MGN-LSTM can accurately predict the temporal evolution of gas saturation and pore pressure in a CO$_2$ injection case with two impermeable faults.
The model exhibits excellent generalizability to mesh configurations, well locations, and permeability fields not included in the training set.
Furthermore, our comparison study shows that MGN-LSTM is more accurate than standard MGN and predicts dynamic fields with a smaller temporal error accumulation thanks to the LSTM cell.
These conclusions have key implications in the field of CO$_2$ storage since accurately capturing uncertainties related to geological structures is critical.
By enabling the construction of surrogates on unstructured meshes, MGN-LSTM has the potential to accelerate the quantification of uncertainties in CO$_2$ injection cases with complex geological structures such as faults and fractures.

Future work in this area should address several topics. 
We observed that using full-graph embeddings to retain recurrent memories in LSTM could require a significant amount of memory, limiting the size of the model that could be trained. 
Future work could explore alternative architecture designs to avoid using full-graph embeddings. 
Furthermore, the model training is currently limited to 2D cases, and scaling up to larger domains remains an open challenge.
Finally, combining data assimilation techniques with the proposed surrogate model is also of interest. 
%

% For now, commented out
\section{Code Availability}
The MGN-LSTM model architecture code and training dataset used in this study can be accessed at~\url{https://github.com/IsaacJu-debug/gnn_flow} upon the publication of this manuscript.

\section{Acknowledgements}
We are grateful for the funding provided by TotalEnergies through the FC-MAELSTROM project. We also acknowledge the Livermore Computing Center and TotalEnergies for providing the computational resources used in this work. We thank Nicola Castelletto for his insights, fruitful discussions, and help with PEBI mesh generation and support in GEOS. Portions of this work were performed during a summer internship of the first author with TotalEnergies in 2022.

\section{Appendix}

%\subsection{Details of high-fidelity (HF) simulation model}
%\label{app:HF_model}
%\todo[inline]{Isaac: Do we need a bit of detail about the relative perm curves?}

%Figure.~\ref{fig:rel_perm} shows the gas–water relative permeability curves used in all HF simulations.

%\begin{figure}[htbp]
%  \centering
%  \includegraphics[trim={0cm 0cm 0cm 0cm},clip, scale=1.0]{figs/appendix/relative_perm_plot.pdf}
%   \caption{Gas–water relative permeability curves used in all HF simulations.}
%  \label{fig:rel_perm}
%\end{figure}

\subsection{Details of MGN-LSTM}
\subsubsection{Encoder}
\label{app:encoder}
This subsection gives the detailed structure for the node level MLP ($\operatorname{mlp}_{v}$) and edge level MLP ($\operatorname{mlp}_{e}$), as shown in Table~\ref{table:encoder}.

\begin{table}[!hpb]
\caption{Details of node and edge encoders}
\begin{center}
\begin{tabular}{ll}
\hline Node MLP $\left(\operatorname{mlp}_{v}\right)$ & Edge MLP $\left(\operatorname{mlp}_{e}\right)$ \\
\hline Input: $x$ & Input: $x$ \\
$x=\operatorname{Linear}(8,100)(x)$ & $x=\operatorname{Linear}(3,100)(x)$ \\
$x=\operatorname{RELU}(x)$ & $x=\operatorname{RELU}(x)$ \\
$x=\operatorname{Linear}(100,100)(x)$ & $x=\operatorname{Linear}(100,100)(x)$ \\
$x=\operatorname{LayerNorm}(100)(x)$ & $x=\operatorname{LayerNorm}(100)(x)$ \\
\hline
\end{tabular}
\label{table:encoder}
\end{center}
\end{table}

\subsubsection{Processor}
\label{app:processor}
As discussed in Section~\ref{subsec:encoder_processor}, the processor is a stack of $m$ graph neural network (GNN) layers, each one performing a round of message passing. The hyperparameter values used for the processor are detailed in Table~\ref{table:processor}.
Except for the input dimension, the structure of $\operatorname{mlp}_{v}$ and $\operatorname{mlp}_{e}$ employed in the processor are the same with the ones used in the encoder (Table~\ref{table:encoder}).

\begin{table}[!hpb]
\caption{Hyperparameters used for processor}
\begin{center}
\begin{tabular}{l|l}
\hline Parameter name & Value \\
\hline Number of GNN layer for the processor & 10 \\
Latent size for the processor & 100 \\
Activation & Relu \\
Type of normalization & Layer normalization \\
Input feature size of Node MLP for one GNN layer & 200\\
Input feature size of Edge MLP for one GNN layer & 300\\
\hline 
\end{tabular}
\label{table:processor}
\end{center}
\end{table}

\subsubsection{Graph Convolutional LSTM cell}
\label{app:lstm}

The GConvLSTM cell employs identical input and output channel dimensions, both set to the processor's latent size of 100. The  filter size of Chebyshev spectral convolutional kernel is set to 8, with symmetric normalization being applied to the graph Laplacian~\cite{seo2018structured}.

\subsubsection{Training hyperparameters}
\label{app:training_hyper}
All hyperparameters employed in the training process are listed in Table~\ref{table:hyper_training}.

\begin{table}[!hpb]
\caption{Hyperparameters used for training}
\begin{center}
\begin{tabular}{l|l}
\hline Parameter name & value \\
\hline Loss function & RMSE \\
Number of epochs & 500 \\
Batch size & 10 \\
Learning rate & 0.001 \\
Weight decay & $5 \mathrm{e}-4$ \\
Optimizer & Adam \\
Optimizer scheduler & cos \\
\hline
\end{tabular}
\end{center}
\label{table:hyper_training}
\end{table}

\subsection{Details of MeshGraphNet (MGN) baseline}
\label{app:MGN}

Unlike MGN-LSTM, where entire sequences of training steps are trained, MGN training involves only minimizing data mismatch between predicted node label and true node label for a single time step.
Moreover, MGN utilizes the same encoder-processor-decoder components and training hyperparameters as those used in MGN-LSTM.

\subsubsection{Noise injection in MGN}
\label{app:noise}
Following the noise injection technique given in~\citep{pfaff2020learning}, we apply a zero-mean Gaussian noise to the field of dynamical quantities, such as gas saturation, during training.
The standard deviation, $s_{n}$, of the noise is a hyperparameter requiring tuning to achieve a stable rollout.
In this work, we have found $s_{n} = 0.05\cdot s $ to be a good choice, where $s$ is the standard deviation of dynamical quantities in the dataset.
We remark that the proposed MGN-LSTM model does not need perturbing corresponding dynamical quantities in the training dataset during the training.

%\newpage
%\section*{References}

\bibliographystyle{elsarticle-num-names}
\bibliography{reference}

\end{document}